\documentclass[3p,times]{elsarticle}

\usepackage{preprint-clean}
\usepackage{amsfonts}       % blackboard math symbols
\usepackage{array, amsmath}
\usepackage{bm}

\usepackage{booktabs}  % For SDS comparison
\usepackage{tabularx}   % for data gen table
\usepackage{graphicx}
\usepackage{subcaption}
\usepackage{float} % Required for [H]

\usepackage{placeins}  % Forces Table locations

\usepackage{tikz} 
\usetikzlibrary{positioning}
\usetikzlibrary{calc}

\usepackage{xcolor}
\usepackage{hyperref}
\hypersetup{
    colorlinks=true,
    linkcolor=blue,  % Sections, figures, tables, equations
    citecolor=blue,  % Citations
    urlcolor=blue    % URLs and email addresses
}

\usepackage{url}

\DeclareMathOperator*{\argmin}{arg\,min}

\volume{00}

\firstpage{1}

\runauth{}

\usepackage{amssymb}

\usepackage[figuresright]{rotating}

\begin{document}

\begin{frontmatter}

\dochead{}

\title{Inverse Synthesis of Shock Time Series from Shock Response Spectra Using Ensembles of Conditional Variational Autoencoders}

\author[lanl]{Adam Watts\corref{cor1}}
\ead{acwatts@lanl.gov}

\author[lanl]{Charles Young}
\ead{cyoung@lanl.gov}

\author[rochester]{Ryan Bowering}
\ead{rbowerin@ur.rochester.edu}

\author[AeroVironment]{Andrew Jeon}
\ead{andrew.jeon@avinc.com}

\author[lanl]{Destry Newton}
\ead{destryn@lanl.gov}

\cortext[cor1]{Corresponding author}

\address[lanl]{Los Alamos National Laboratory,
P.O. Box 1663, Los Alamos, NM 87545, USA}

\address[rochester]{University of Rochester,
Rochester, NY, USA}

\address[AeroVironment]{241 18th Street South, Suite 650, Arlington, VA 22202}

\begin{abstract}
The shock response spectrum (SRS) characterizes the peak responses of single-degree-of-freedom systems to transient acceleration. Because the mapping from an acceleration time series (waveform) to its SRS is many-to-one, inverse synthesis seeks candidate waveforms consistent with a prescribed spectrum. We extend an earlier conditional variational autoencoder (CVAE) framework with a recurrent decoder that predicts multiscale wavelet coefficients and reconstructs waveforms through an inverse discrete wavelet transform. Individual models exhibit limited variability under latent sampling, motivating an ensemble of independently trained models spanning different mother wavelets and decomposition levels. Each member produces a separate candidate waveform for the same target SRS. Evaluation on four held-out benchmark datasets shows lower mean and median spectral errors for the illustrated wavelet model than for the earlier direct time-domain CVAE and a sum-of-decayed-sinusoids (SDS) baseline. A conditioning study examines frequency-grid and encoding choices. Comparisons with published inverse-SRS methods demonstrate superior performance across four holdout benchmark datasets—two operational shock datasets, one earthquake dataset, and one synthetic shock dataset—and competitive performance on synthetic targets reported by Brake~\cite{brake_inverse_2011}. Single-model inference requires approximately 0.1--0.6~ms per waveform during batched inference compared with 5--8~s for SDS, excluding offline training. This low inference cost enables rapid generation of distinct SRS-compatible candidates waveforms across the ensemble, although their physical suitability requires assessment beyond spectral agreement.

\end{abstract}


\end{frontmatter}

\section{Introduction}

A mechanical shock is a short-duration transient characterized by a rapid change in velocity and typically high acceleration~\cite{lalanne_mechanical_2014}. Building on Biot's study of transient single-degree-of-freedom (SDOF) responses~\cite{biot_transient_1932}, the shock response spectrum (SRS) characterizes a shock through the peak responses of oscillators with different natural frequencies and a specified damping ratio. It is widely used in equipment design and qualification, including the environmental testing procedures described in MIL-STD-810G~\cite{united_states_department_of_defense_mil-std-810g_2008}. This work considers the maximax acceleration SRS, which retains each oscillator's largest absolute acceleration response over time.

Laboratory reproduction of a target SRS requires an acceleration time series (waveform), but the spectrum does not uniquely determine that waveform. Environmental specifications may also comprise frequency-wise statistical envelopes or tolerance bounds assembled from multiple measured shocks. These targets do not preserve the timing, phase, sign, or cross-frequency dependence of any individual event. Inverse SRS synthesis therefore seeks one or more waveforms whose spectra match the target within prescribed tolerances, rather than recovering a unique original signal \cite{lalanne_mechanical_2014}. Multiple candidates are useful because a spectrally compatible waveform may be infeasible to reproduce within the operating limits of the test equipment.

Conventional methods commonly optimize a finite sum of exponentially decaying sinusoids for each target~\cite{smallwood_matching_1974}. Extensions incorporate additional basis families and genetic-algorithm refinement~\cite{brake_inverse_2011}, temporal-moment constraints~\cite{cap_methodology_1997}, or Monte Carlo optimization~\cite{cap_monte_carlo_2018}. Their performance depends on the waveform parameterization and optimization procedure; repeated optimization can be expensive, and a prescribed finite basis restricts the available waveform morphologies.

Time-series foundation models have been investigated for forecasting, imputation, classification, and anomaly detection, among other tasks~\cite{kottapalli2025foundation}. The inverse SRS problem considered here requires synthesizing a complete acceleration time history conditioned on a prescribed response spectrum, without an observed waveform or temporal context supplied at inference.

Learning-based generative models offer an alternative to per-target optimization by learning a mapping from prescribed spectra to compatible waveforms. Within these models, the signal representation provides a further design choice. For example, Yuan and Qiao~\cite{yuan_diffusion-ts_2024} incorporated Fourier-based representations and a Fourier-domain loss into Diffusion-TS. Wavelets represent signals across both scale and temporal location, making them relevant to localized transient structure. Wavelet-based generative methods have also been investigated for images and time series~\cite{phung2023wavelet, wang2025waveletdiff}.

Our earlier preprint introduced a CVAE framework for inverse SRS reconstruction, a synthetic shock generation procedure, and four benchmark datasets~\cite{watts2026inverse}. The present study extends that framework with a recurrent wavelet decoder and an ensemble of independently trained models spanning different mother wavelets and decomposition levels. Individual models produced nearly identical waveforms under repeated latent sampling, behavior consistent with posterior collapse~\cite{lucas2019collapse}; alternative candidates therefore arise primarily across models. Using the previously introduced benchmarks, we compare the wavelet decoder with the earlier direct time-domain CVAE and SDS. We also examine SRS conditioning configurations and compare spectral agreement with published inverse-SRS results.

\subsection{Contributions}

The principal contributions of this study are:
\begin{itemize}
    \item \textbf{Recurrent wavelet decoding:}
    A decoder predicts multiscale wavelet coefficients sequentially from coarse to fine and reconstructs acceleration through an inverse discrete wavelet transform (IDWT).

    \item \textbf{Ensemble generation of candidate solutions:}
    Independently trained models with different mother wavelets and decomposition levels provide distinct waveforms for a common target SRS, addressing the limited variability observed under latent sampling within individual models.

    \item \textbf{SRS conditioning study:}
    A controlled comparison evaluates how frequency-grid selection and concatenation versus Hadamard encoding affect reconstruction accuracy.

    \item \textbf{Comparative evaluation:}
    Experiments on the four previously introduced benchmarks assess the wavelet decoder against the earlier direct time-domain CVAE and SDS. Additional comparisons with published inverse-SRS reconstructions and runtime measurements characterize spectral agreement and computational cost.
\end{itemize}

\section{Background and Foundational Methods}

\subsection{The Shock Response Spectrum (SRS)}

For an input acceleration $\ddot{x}[n]$, where $n$ is the integer-valued discrete-time sample index, let $r_i[n]$ denote the acceleration response of an SDOF oscillator with natural frequency $\mathfrak{f}_i$ and damping ratio $\zeta$. Its maximax SRS ordinate is
\[
\mathfrak{s}_i=\max_n |r_i[n]|,
\qquad
\bm{\mathfrak{s}}
=\operatorname{SRS}\!\left\{\ddot{x}[n],\bm{\mathfrak{f}};\zeta\right\},
\]
where $\bm{\mathfrak{f}}=(\mathfrak{f}_1,\ldots,\mathfrak{f}_F)$ contains the evaluation natural frequencies. All SRS calculations use a damping ratio of $\zeta=0.03$, which is omitted from subsequent notation. Each acceleration record is extended with a zero-valued tail padding to capture residual oscillator responses after the input ends. The tail duration is approximately one-third of the period of the lowest-frequency oscillator. The Fraktur symbol $\mathfrak{f}$ distinguishes SDOF evaluation frequencies from component frequencies used to generate synthetic shocks. The oscillator formulation, padding, and numerical SRS implementations are described in our earlier work~\cite{watts2026inverse}.

\subsection{Inverse Problem Formulation}
\label{sec:inverse_problem}

Given a target spectrum $\bm{\mathfrak{s}}_{\mathrm{target}}$, inverse synthesis seeks acceleration waveforms whose calculated SRS lies within an allowable error $\epsilon$. For a prescribed spectral distance $d$, the admissible set is
\[
\mathcal{A}\!\left(\bm{\mathfrak{s}}_{\mathrm{target}};\epsilon\right)
=
\left\{
\hat{x}[n]\;\middle|\;
d\!\left(
\operatorname{SRS}\!\left\{\hat{x}[n],\bm{\mathfrak{f}}\right\},
\bm{\mathfrak{s}}_{\mathrm{target}}
\right)\leq\epsilon
\right\}.
\]
Here, $\hat{x}[n]$ denotes acceleration directly, with the double-dot notation omitted for simplicity. Because distinct waveforms can satisfy the same spectral constraint, the inverse is generally set-valued. The objective is to generate candidate members of this set; agreement with the target SRS alone does not establish their suitability for a particular physical test.

\subsection{Conventional Inverse SRS Methods}

A common approach to reconstructing a candidate acceleration time series
$\hat{x}[n]$ from a given SRS curve is to formulate the inverse problem as an
optimization task using exponentially decaying sinusoids as basis
functions~\cite{smallwood_matching_1974}. A sum of exponentially decaying
sinusoids can be expressed as
\begin{equation}
\label{EQ:SumDecayedSinesDiscrete}
x_{\mathrm{SDS}}[n;\boldsymbol{\theta}]
=
\sum_{m=1}^{M}
A_m e^{-\lambda_m nT_s}
\sin\!\left(2\pi f_m nT_s+\phi_m\right),
\end{equation}
where each of the $M$ components is parameterized by an amplitude $A_m$,
decay rate $\lambda_m$, frequency $f_m$ in Hz, and phase $\phi_m$. The
sampling period is $T_s=1/f_s$, where $f_s$ is the sampling frequency. The parameter vector $\boldsymbol{\theta}$ collects these $4M$ component parameters.

A typical distance metric for comparing the reconstructed and target spectra
is the root-mean-square logarithmic error (RMSLE), computed in base-10
logarithmic space:
\begin{equation}
\label{EQ:SRSRMSLE}
d_{\mathrm{RMSLE}}
\left(
\bm{\mathfrak{s}},
\bm{\mathfrak{s}}_{\mathrm{target}}
\right)
=
\sqrt{
\frac{1}{F}
\sum_{j=1}^{F}
\left[
\log_{10}
\left(
\frac{\mathfrak{s}_j}
{\mathfrak{s}_{\mathrm{target},j}}
\right)
\right]^2
},
\end{equation}
where $F$ is the number of SRS evaluation frequencies. The SDS optimization
problem can then be written as
\begin{equation}
\label{EQ:OptimizeSumDecayedSines}
\hat{\boldsymbol{\theta}}_{\mathrm{SDS}}
=
\argmin_{\boldsymbol{\theta}\in\Theta}
d_{\mathrm{RMSLE}}
\left(
\operatorname{SRS}
\left(
x_{\mathrm{SDS}}[n;\boldsymbol{\theta}]
\right),
\bm{\mathfrak{s}}_{\mathrm{target}}
\right).
\end{equation}
The optimized parameters define the reconstructed waveform $\hat{x}_{\mathrm{SDS}}[n]$. Related methods augment spectral matching with temporal moments~\cite{cap_methodology_1997} or expand the basis to include impulses, sine functions, and modified Morlet wavelets~\cite{brake_inverse_2011}. The present benchmarks distinguish the SDS baseline from Brake's published reconstructions and their genetic-algorithm refinements.

The SDS baseline used in the present study employs the same in-house MATLAB implementation as our earlier study \cite{watts2026inverse}.  It used 12 tones per octave, a damping ratio of 0.03, and two sweep iterations. The reported accuracy and runtime apply to this configuration. Alternative SDS implementations are available, including SDynPy~\cite{sdynpy_sds}.

\section{Proposed Method}
\subsection{Conditional Variational Autoencoders}

A variational autoencoder (VAE) couples an encoder that infers a latent representation with a decoder that generates observations from that representation~\cite{kingma_autoencoding_2014}. A conditional VAE supplies additional information to both networks~\cite{sohn_learning_2015}. Here, $\mathbf{x}\in\mathbb{R}^{N}$ denotes the reference acceleration time series and $\mathbf{c}$ its SRS conditioning information. The encoder approximates the posterior $q_\phi(\mathbf{z}\mid\mathbf{x},\mathbf{c})$, while the decoder parameterizes $p_\theta(\mathbf{x}\mid\mathbf{z},\mathbf{c})$. The prior is $p(\mathbf{z})=\mathcal{N}(\mathbf{0},\mathbf{I})$, independent of the conditioning information.

During training, latent samples are drawn from the encoder's approximate posterior using the reparameterization trick. During inference, the encoder is omitted: a latent sample from the prior and the target SRS conditioning are supplied to the decoder to generate $\hat{\mathbf{x}}$. Figure~\ref{fig:cvae_architecture_wavelet_recurrent} shows the architecture. Although the latent formulation permits stochastic generation, the trained models showed little waveform variability for a fixed target; the ensemble described in Section~\ref{sec:Multi-Model Wavelet Ensemble} supplies alternative candidates through independently trained models.

Preliminary experiments suggested that wavelet representations were more beneficial in the decoder than in the encoder. A systematic comparison of wavelet-based encoders was not conducted. The present architecture therefore combines a conventional encoder with a recurrent wavelet decoder that reconstructs the acceleration time series across multiple scales.

\subsection{SRS--Frequency Conditioning}
\label{sec:srs_frequency_conditioning}

We compare concatenation and Hadamard representations of the amplitude-normalized target SRS $\bm{\mathfrak{s}}_{\mathrm{norm}}$, defined in \ref{sec:Normalization}, and its natural frequencies $\bm{\mathfrak{f}}$.

With numerical floor $\epsilon>0$, define the log-transformed
SRS ordinates as
\begin{equation}
\tilde{\mathfrak{s}}_j
=\log_{10}\!\left[\max\!\left(
\mathfrak{s}_{\mathrm{norm},j},\epsilon\right)\right].
\end{equation}
Concatenation conditioning uses
\begin{equation}
\mathbf{c}_{\mathrm C}
=\begin{bmatrix}
\tilde{\bm{\mathfrak{s}}}\\
\tilde{\bm{\mathfrak{f}}}
\end{bmatrix}
\in\mathbb{R}^{2F},
\end{equation}
where the frequency entries are log-transformed and normalized as
\begin{equation}
\tilde{\mathfrak{f}}_j
=2\frac{\log_{10}(\mathfrak{f}_j)
-\log_{10}(\mathfrak{f}_{\min})}
{\log_{10}(\mathfrak{f}_{\max})
-\log_{10}(\mathfrak{f}_{\min})}-1.
\end{equation}
Here, $\mathfrak{f}_{\min}$ and $\mathfrak{f}_{\max}$ denote
the frequency bounds mapped to $-1$ and $1$, respectively.

Hadamard conditioning instead uses
\begin{equation}
\mathbf{c}_{\mathrm H}
=\log_{10}\!\left[
\max\!\left(
\bm{\mathfrak{s}}_{\mathrm{norm}}\odot\bm{\mathfrak{f}},
\epsilon\right)\right]
\in\mathbb{R}^{F},
\end{equation}
with elementwise operations and no frequency rescaling.
The model library uses Hadamard conditioning;
\ref{sec:srs_conditioning_study} compares the encoding
and frequency-grid choices.

\subsection{Encoder}

The encoder architecture is unchanged from our earlier study \cite{watts2026inverse}. It concatenates the acceleration time series with its SRS embedding and applies four one-dimensional convolutional layers with max-pooling and GELU activations, followed by flattening and two fully connected layers. Separate linear heads predict the mean and log-variance of a 100-dimensional Gaussian latent distribution, sampled through reparameterization.

\subsection{Recurrent Wavelet Decoder}

A discrete wavelet transform recursively applies low-pass and high-pass filters with downsampling, producing coarse approximation and finer detail coefficients~\cite{mallat1989multiresolution}. Mother wavelets differ in support, smoothness, and symmetry; their family, order, and decomposition level are therefore model hyperparameters in the proposed framework. The decoder predicts coefficient sets from coarse to fine and reconstructs the waveform through the IDWT. Coefficient dimensions depend on the signal length, wavelet, level, and boundary convention and are fixed when the model is instantiated. The PyTorch Wavelet Toolbox (PTWT) ~\cite{wolter2024ptwt} provides GPU-compatible differentiable transforms, enabling end-to-end training through the IDWT.

A two-layer MLP with GELU activations maps the latent vector and SRS conditioning to a global representation $\mathbf{g}\in\mathbb{R}^{512}$. This initializes the recurrent state $\mathbf{h}_0=\mathbf{g}$. For coefficient set $k$, a BandBlock combines the preceding state with a learned embedding $\mathbf{e}_k\in\mathbb{R}^{512}$ and the fixed global representation:
\begin{equation}
\begin{aligned}
\mathbf{s}_k&=\operatorname{LN}(\mathbf{h}_{k-1}+\mathbf{e}_k),\\
\mathbf{u}_k&=\operatorname{concat}[\mathbf{s}_k,\operatorname{LN}(\mathbf{g})]\in\mathbb{R}^{1024},\\
\boldsymbol{\Delta}_k&=W_{2,k}\operatorname{GELU}(W_{1,k}\mathbf{u}_k+\mathbf{b}_{1,k})+\mathbf{b}_{2,k},\\
\mathbf{h}_k&=\operatorname{GELU}\!\left[\operatorname{LN}(\mathbf{s}_k+\boldsymbol{\Delta}_k)\right],\\
\hat{\mathbf{c}}_k&=W_k^{(c)}\mathbf{h}_k+\mathbf{b}_k^{(c)}.
\end{aligned}
\end{equation}
Here, LN denotes layer normalization. Each linear output head has the dimension of its coefficient set, while the recurrent state passes coarse-scale information to subsequent finer scales.

\begin{figure}[htbp]
\centering
\includegraphics[width=\textwidth]{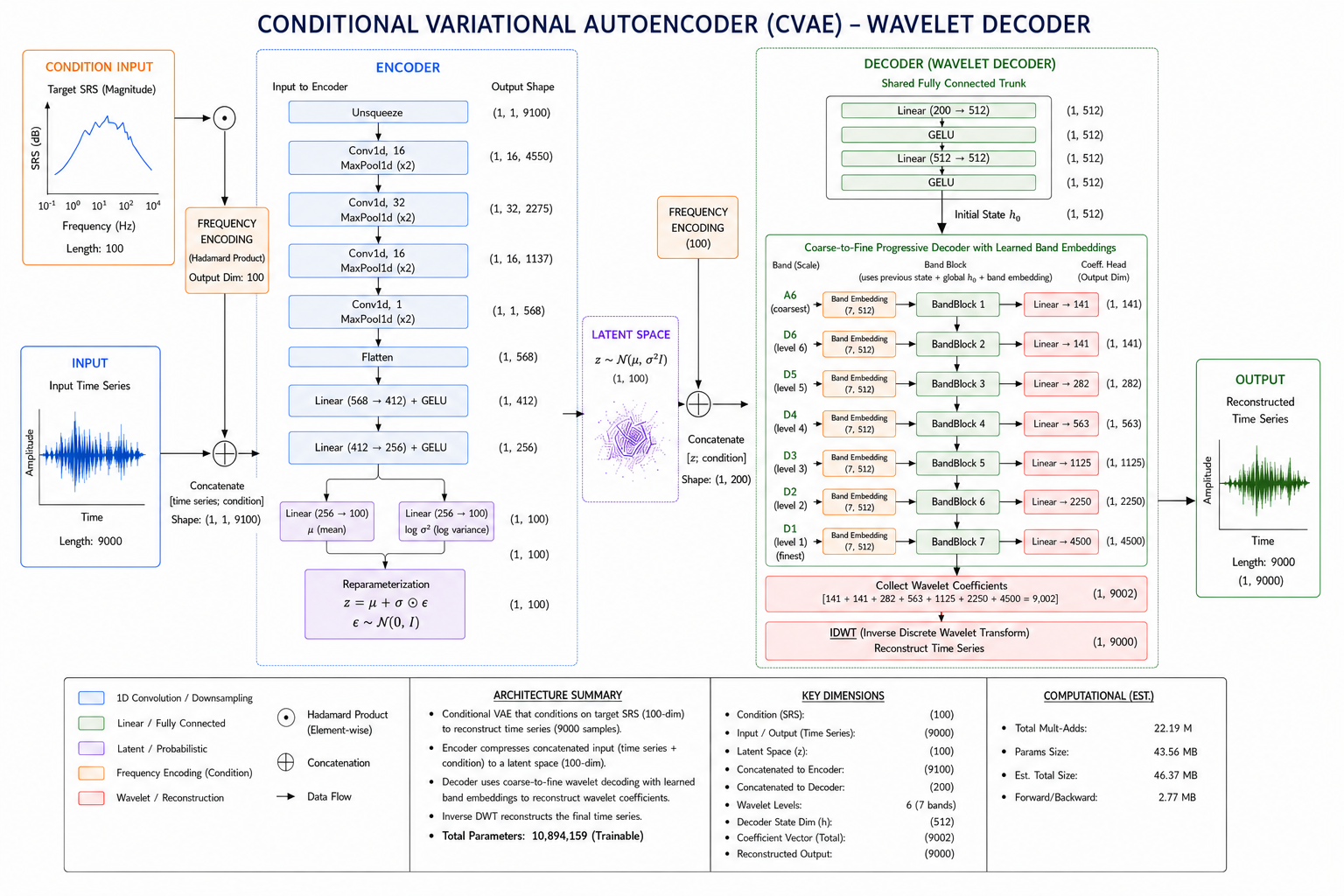}
\caption{CVAE with recurrent wavelet decoder, illustrated for a six-level wavelet decomposition. BandBlocks predict coefficient sets from coarse to fine; the IDWT reconstructs the 9000-sample acceleration time series.}
\label{fig:cvae_architecture_wavelet_recurrent}
\end{figure}

For the six-level wavelet decomposition example in Figure~\ref{fig:cvae_architecture_wavelet_recurrent}, the ordered coefficient sets $(A_6,D_6,D_5,D_4,D_3,D_2,D_1)$ have dimensions $(141,141,282,563,1125,2250,4500)$. The decoder produces
\begin{equation}
\widehat{\mathcal C}_6=(\hat{\mathbf a}_6,\hat{\mathbf d}_6,\hat{\mathbf d}_5,\hat{\mathbf d}_4,\hat{\mathbf d}_3,\hat{\mathbf d}_2,\hat{\mathbf d}_1),
\label{EQ:PredictedWaveletCoefficients}
\end{equation}
and reconstructs
\begin{equation}
\hat{\mathbf x}=\operatorname{IDWT}_{\psi,\beta}(\widehat{\mathcal C}_6)\in\mathbb{R}^{9000},
\label{EQ:WaveletDecoderIDWT}
\end{equation}
where $\psi$ and $\beta$ specify the mother wavelet and boundary convention. The 9002 predicted coefficients yield 9000 output samples under the selected boundary-handling convention; the complete DWT coefficient count is at least the signal length and may exceed it.

\subsection{Loss Function}\label{sec:Loss}

Training follows the composite objective introduced in our earlier
CVAE framework~\cite{watts2026inverse}, combining waveform-shape,
time-series, power spectral density (PSD), shock response spectrum
(SRS), and Kullback--Leibler (KL) divergence terms:
\begin{equation}
\mathcal{L}_{\mathrm{total}}
=
\lambda_{\mathrm{shape}}\mathcal{L}_{\mathrm{shape}}
+
\lambda_{\mathrm{TS}}\mathcal{L}_{\mathrm{TS}}
+
\lambda_{\mathrm{PSD}}\mathcal{L}_{\mathrm{PSD}}
+
\lambda_{\mathrm{SRS}}\mathcal{L}_{\mathrm{SRS}}
+
\lambda_{\mathrm{KL}}\mathcal{L}_{\mathrm{KL}}.
\label{eq:ltotal}
\end{equation}

The waveform-shape term compares absolute SDOF acceleration response histories near their respective maxima. Both the earlier direct time-domain CVAE and the present wavelet CVAE used an accelerated implementation that gathers local windows aligned at the maximum absolute responses, with soft peak locations used for window sizing and Gaussian weighting. This implementation was not detailed in the earlier paper and is documented in \ref{app:loss_implementation}, together with the remaining loss definitions and coefficients.

\subsection{Multi-Model Wavelet Ensemble}
\label{sec:Multi-Model Wavelet Ensemble}

Individual models produced nearly identical waveforms under repeated latent sampling for a fixed target, consistent with posterior collapse. We therefore trained 104 models spanning biorthogonal (\texttt{bior}), Symlet (\texttt{sym}), and Daubechies (\texttt{db}) wavelets at multiple decomposition levels.  Candidate decomposition levels ranged from 2 to 11, subject to the maximum useful level returned by \texttt{pywt.dwt\_max\_level} for a 9000-sample signal and each wavelet's decomposition filter length. Consequently, wavelets with longer filters had lower maximum candidate levels.

For $K$ independently trained CVAE models, the candidate set at inference is
\begin{equation}
\widehat{\mathcal X}_{\mathrm{ens}}
=\{\hat{x}_m[n]=g_{\boldsymbol{\theta}_m}(\mathbf z_m,\mathbf c_m)\}_{m=1}^{K},
\qquad \mathbf z_m\sim\mathcal N(\mathbf 0,\mathbf I),
\end{equation}
where $g_{\boldsymbol{\theta}_m}$ is the decoder of model $m$, trained jointly with its encoder, and $\mathbf c_m$ encodes the common target SRS and its frequencies for that model. Each waveform is retained separately for assessment against spectral tolerances and physical test constraints; waveforms are not averaged. Variability arises primarily across models.

\section{Datasets and Preprocessing}

\subsection{Synthetic Data Generation}

Synthetic waveforms were generated using the framework introduced in our earlier work~\cite{watts2026inverse}. Preliminary experiments favored combinations of exponentially decaying sinusoids and modified Morlet-like pulses for generalization to measured shocks. The framework also supports radial basis functions, sawtooth functions, and mother wavelets available through PyWavelets~\cite{lee2019pywavelets}. The analytical components define the synthetic training distribution; inference predicts wavelet coefficients rather than optimizing these generator components.

All synthetic training waveforms contain $N=9000$ time samples at $f_s=32{,}768$~Hz, corresponding to approximately 0.2747~s. Each synthetic waveform combines one to ten randomly parameterized components and additive noise. Successive components either share the preceding temporal offset or receive an independently sampled offset, producing both clustered and separated transients. Basis definitions, parameter distributions, and placement rules are provided in \ref{sec:synthetic_generation_appendix}; experiment-specific frequency-range changes are stated with the corresponding comparisons.

\subsection{Training Data and Optimizer}

The model-development dataset contained 400{,}000 synthetic and 99{,}682 measured shock waveforms, for a total of 499{,}682 examples. Each waveform $\mathbf{x}_j\in\mathbb{R}^{N}$ was paired with its computed SRS, $\bm{\mathfrak{s}}_j=\operatorname{SRS}\{\mathbf{x}_j,\boldsymbol{\mathfrak{f}}\}$. A 99\%/1\% split produced an internal training set of 494{,}685 examples and a held-out set of 4{,}997 examples. Training used the AdamW optimizer with a batch size of 128 and a learning rate of $1.43244\times10^{-4}$. All models were trained for 150 epochs, and the final checkpoint was used for evaluation. The four independent benchmark datasets described below were separate from this model-development split.

\subsection{Evaluation Datasets}

Final evaluation used the four previously introduced benchmark datasets, denoted $\mathcal{D}_{\mathrm{test}}^{j}$ for $j\in\{A,B,C,D\}$~\cite{watts2026inverse}. Datasets A and B contain operational shock recordings, with 963 and 890 examples, respectively. Their evaluated versions retain the original physical amplitude scaling; normalized and anonymized versions are publicly released, while the original recordings are withheld because they contain identifying test signatures. Dataset C contains 815 examples extracted from earthquake recordings, and dataset D contains 1000 synthetic examples combining exponentially decaying sinusoids, modified Morlet-like pulses, sawtooth functions, and radial basis functions.

Datasets A, B, and D were natively sampled at 32.768~kHz. The earthquake recordings, originally 9--41~s long and sampled at lower rates, were resampled to 32.768~kHz. Records in A, B, and C were then windowed to 9000 samples (approximately 0.2747~s), positioning the maximum amplitude at approximately 5\% of the window. The earthquake evaluation therefore concerns these short excerpts. The common sampling rate and length match the model's input and output dimensions; all target spectra were calculated from the resulting evaluation time series.

\subsection{Normalization}\label{sec:Normalization}

Each acceleration time series and its SRS are divided by the largest SRS ordinate. This paired normalization follows from the homogeneity of the maximax SRS:
\begin{equation}
\operatorname{SRS}\{c\mathbf{x},\boldsymbol{\mathfrak{f}}\}
=|c|\operatorname{SRS}\{\mathbf{x},\boldsymbol{\mathfrak{f}}\}.
\label{eq:srs_scaling}
\end{equation}
For a nonzero maximax spectrum, define $a=\|\bm{\mathfrak{s}}\|_\infty$.
The normalized pair is
\begin{equation}
\bm{\mathfrak{s}}_{\mathrm{norm}}=\frac{\bm{\mathfrak{s}}}{a},
\qquad
\mathbf{x}_{\mathrm{norm}}=\frac{\mathbf{x}}{a},
\label{eq:paired_normalization}
\end{equation}
so the largest normalized SRS ordinate is one. At inference, the target SRS is normalized in the same way and the generated waveform is restored to physical units using
\begin{equation}
\hat{\mathbf{x}}_{\mathrm{phys}}=a\hat{\mathbf{x}}_{\mathrm{norm}}.
\label{eq:inverse_srs_normalization}
\end{equation}

\section{Results}

\subsection{Evaluation Procedure and Performance Metrics}

Spectral agreement is assessed using the per-sample RMSLE defined in Eq.~\eqref{EQ:SRSRMSLE} and summarizes spectrum agreement across each SRS \textit{sample}. We also report the fraction of benchmark samples for which the CVAE achieves lower RMSLE than SDS for a representative model from our ensemble, Figure \ref{fig:Holdout_RMSLE_ABCD}. 

The absolute per-frequency dB error:
\begin{equation}
e_j =
\left|
20\log_{10}\!\left(
\frac{\mathfrak{s}_j}{\mathfrak{s}_{\mathrm{target},j}}
\right)
\right|
\quad\text{(dB)},
\label{eq:db_error}
\end{equation} 
measures the error magnitude at each \textit{individual} natural frequency. We pool these errors across benchmark samples and evaluation frequencies and report the empirical fractions at or below 1~dB and 3~dB. These fractions describe individual SRS points, not the fraction of complete spectra satisfying the tolerance at every frequency. Figure~\ref{fig:Holdout_dbEror_ABCD} shows the pooled per-frequency
absolute dB error distributions for the same representative model and SDS.

The normalized correlation coefficient is insufficient as a standalone measure of SRS reconstruction accuracy because it is insensitive to uniform amplitude scaling and gives greater weight to high-amplitude spectral regions. Consequently, strong correlation can coexist with substantial magnitude errors, particularly in lower-amplitude regions of spectra spanning several orders of magnitude.

Most of our comparison is against the SDS methodology because it is a widely utilized technique used to synthesis time-domain pulses from an SRS \cite{ranieri2023comparative}.

\subsection{Representative Ensemble Reconstructions}
\label{sec:Representative}

\begin{figure*}[htbp]
    \centering
    \includegraphics[width=0.98\textwidth]{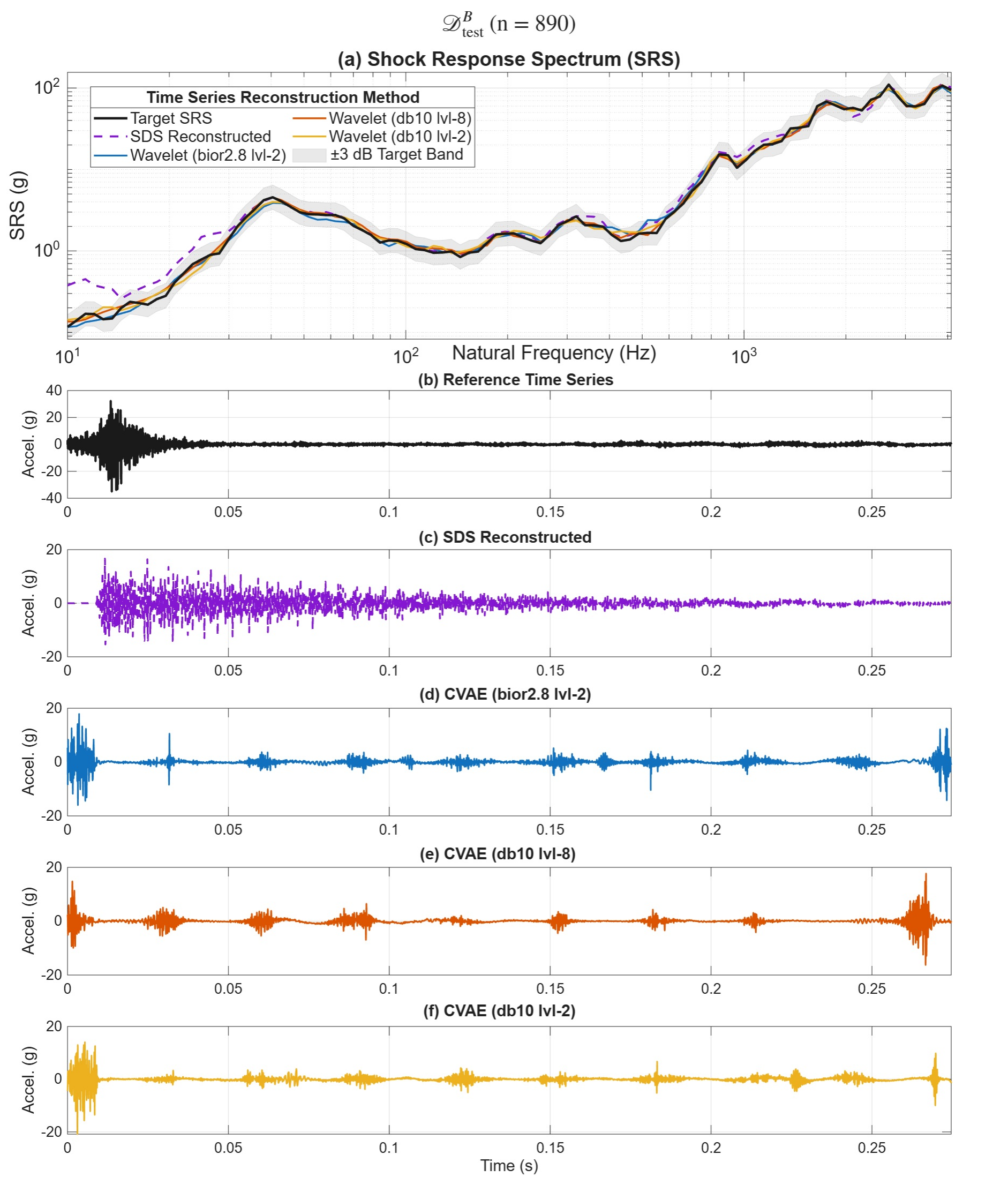}
    \caption{Representative reconstructions for one example from
    $\mathcal{D}_{\mathrm{test}}^{B}$ ($n=890$):
    (a) target and reconstructed SRS, with the shaded region indicating
    the $\pm 3$~dB target band;
    (b) reference acceleration time series used to compute the target SRS;
    (c) SDS reconstruction; and
    (d)--(f) reconstructions from three CVAE ensemble members.
    The reconstructions exhibit similar SRS responses despite
    distinct amplitudes and temporal structures.
    The reference time series is plotted over $\pm 40$~g,
    whereas the reconstructions are plotted over $\pm 20$~g.}
    \label{fig:representative-ensemble}
\end{figure*}

Figure~\ref{fig:representative-ensemble} compares a reference acceleration time series from $\mathcal{D}_{\mathrm{test}}^{B}$, its SDS reconstruction, and three ensemble candidates. The target SRS is evaluated at 100 logarithmically spaced natural frequencies from 10 to 4096~Hz. These examples illustrate between-model waveform variability for the selected target; they do not quantify ensemble diversity across the benchmark datasets. Distinct amplitudes and temporal structures produce similar spectra, illustrating the inverse problem's nonuniqueness. Quantifying candidate diversity and physical suitability requires additional temporal, energy, and frequency-domain measures.

\subsection{Performance on Held-Out Benchmark Datasets}
\label{sec:performance}

The models in the library were trained using Hadamard conditioning before the conditioning study reported in \ref{sec:srs_conditioning_study}, which favored concatenation on the evaluated fixed grids. The effect of retraining the full library with concatenation has not been evaluated. 

The Daubechies~10 (\texttt{db10}), level~8 model was selected for detailed presentation because it showed good reconstruction performance in a qualitative review of the model library, without being consistently the best- or worst-performing configuration. Figure~\ref{fig:representative-ensemble} includes a representative waveform generated by this model configuration. Benchmark results for all evaluated configurations are provided in \ref{app:heldout_rmsle_results} and \ref{app:heldout_db_results}.

Table~\ref{tab:direct_wavelet_comparison} compares this  \texttt{db10}, level~8 model with the earlier direct time-domain CVAE~\cite{watts2026inverse} and SDS. Both CVAE configurations used the same training data, conditioning, loss formulation---including the accelerated waveform-shape implementation---and training settings, with only the decoder architecture changed.

\begin{table}[htbp]
    \centering
    \caption{Descriptive statistics of per-sample RMSLE across the
    four held-out benchmark datasets. SDS and Direct CVAE results
    are reproduced from our earlier work~\cite{watts2026inverse}.
    Wavelet CVAE denotes the \texttt{db10}, level~8 model.
    Boldface identifies the lowest mean and median within each
    dataset. Lower RMSLE indicates closer agreement with the
    target SRS.}
    \label{tab:direct_wavelet_comparison}
    \resizebox{\textwidth}{!}{%
    \begin{tabular}{llrrrrrrr}
        \hline
        Holdout & Method & Mean & Median & Std.\ Dev. &
        Min & Max & 0.025 Quant. & 0.975 Quant. \\
        \hline
        A & SDS
          & 0.206 & 0.176 & 0.110 & 0.051 & 0.894 & 0.082 & 0.489 \\
          & Direct CVAE
          & 0.095 & 0.090 & 0.024 & 0.049 & 0.211 & 0.060 & 0.156 \\
          & Wavelet CVAE
          & \textbf{0.065} & \textbf{0.058}
          & 0.033 & 0.024 & 0.368 & 0.031 & 0.148 \\
        \hline
        B & SDS
          & 0.231 & 0.200 & 0.112 & 0.088 & 0.925 & 0.110 & 0.546 \\
          & Direct CVAE
          & 0.136 & 0.125 & 0.047 & 0.063 & 0.418 & 0.079 & 0.266 \\
          & Wavelet CVAE
          & \textbf{0.094} & \textbf{0.086}
          & 0.029 & 0.045 & 0.376 & 0.051 & 0.161 \\
        \hline
        C & SDS
          & 0.193 & 0.154 & 0.123 & 0.038 & 0.846 & 0.067 & 0.544 \\
          & Direct CVAE
          & 0.113 & 0.109 & 0.029 & 0.059 & 0.269 & 0.072 & 0.186 \\
          & Wavelet CVAE
          & \textbf{0.087} & \textbf{0.082}
          & 0.030 & 0.033 & 0.274 & 0.044 & 0.160 \\
        \hline
        D & SDS
          & 0.152 & 0.131 & 0.081 & 0.046 & 0.907 & 0.068 & 0.369 \\
          & Direct CVAE
          & 0.095 & 0.083 & 0.042 & 0.028 & 0.437 & 0.044 & 0.201 \\
          & Wavelet CVAE
          & \textbf{0.062} & \textbf{0.048}
          & 0.054 & 0.019 & 0.837 & 0.022 & 0.181 \\
        \hline
    \end{tabular}%
    }
\end{table}

\begin{figure}[H]
    \centering
    \begin{subfigure}{0.48\textwidth}
        \centering
        \includegraphics[width=\linewidth]{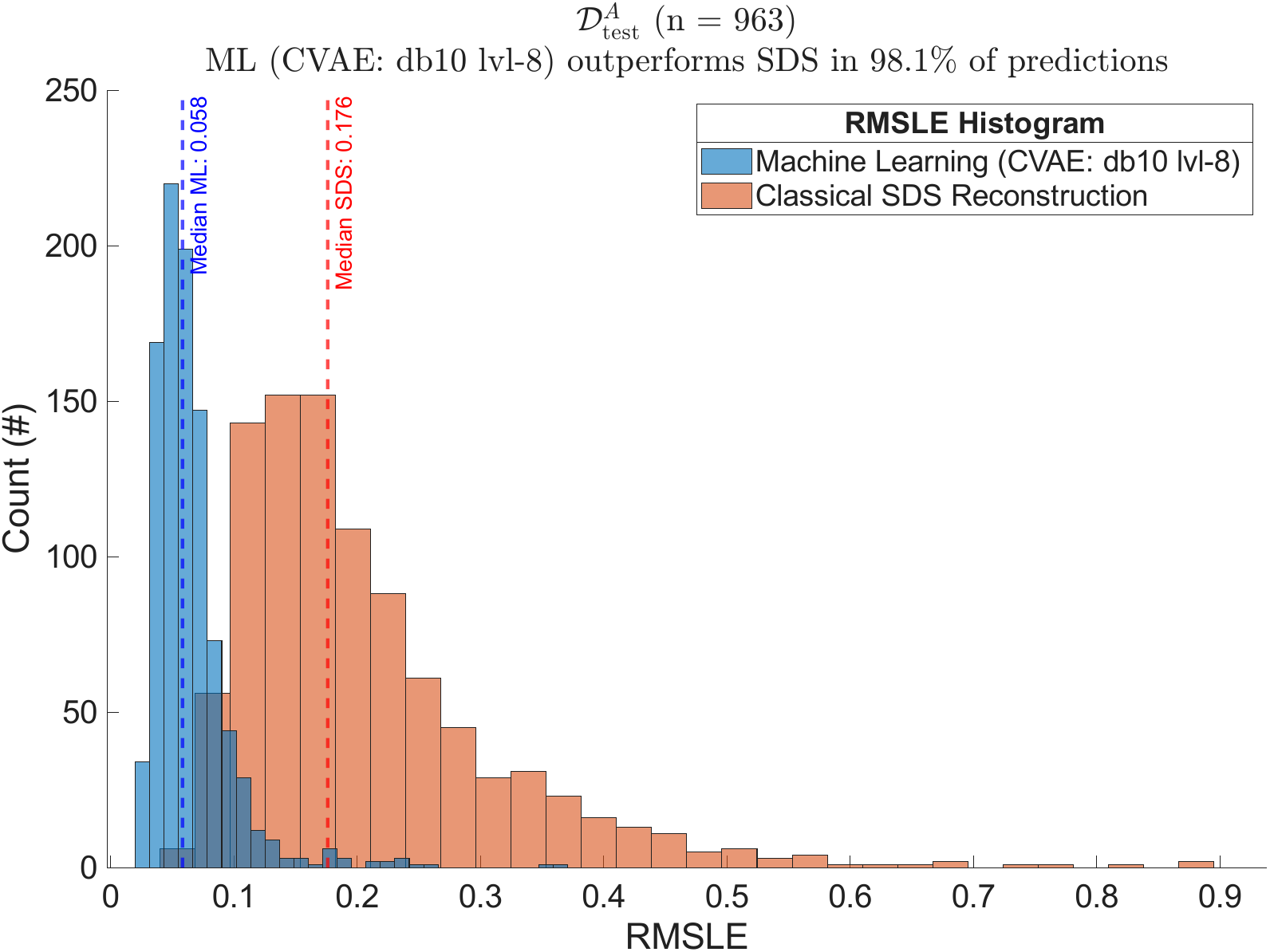}
    \end{subfigure}\hfill%
    \begin{subfigure}{0.48\textwidth}
        \centering
        \includegraphics[width=\linewidth]{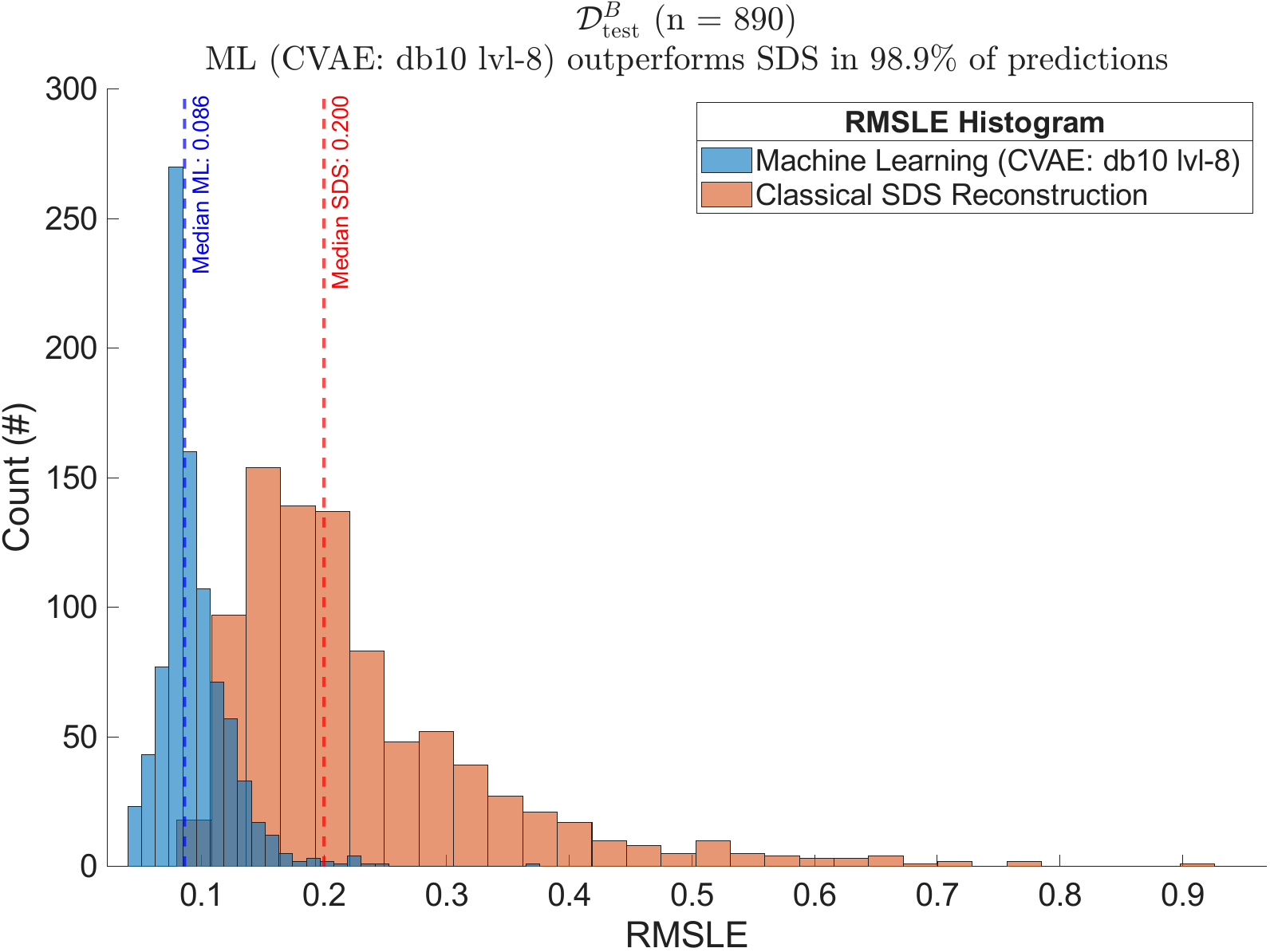}
    \end{subfigure}\par
    \vspace{1em}

    \begin{subfigure}{0.48\textwidth}
        \centering
        \includegraphics[width=\linewidth]{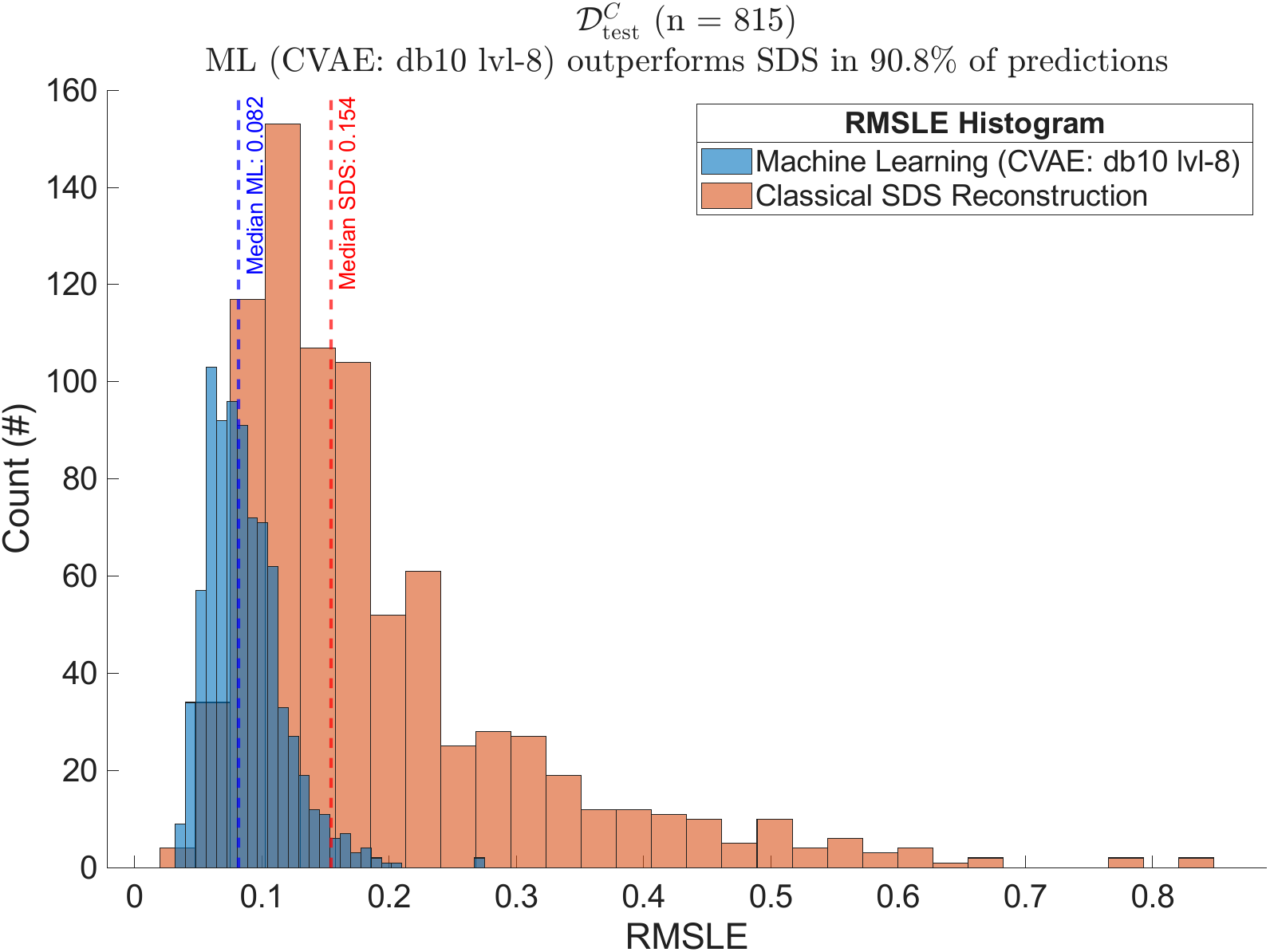}
    \end{subfigure}\hfill%
    \begin{subfigure}{0.48\textwidth}
        \centering
        \includegraphics[width=\linewidth]{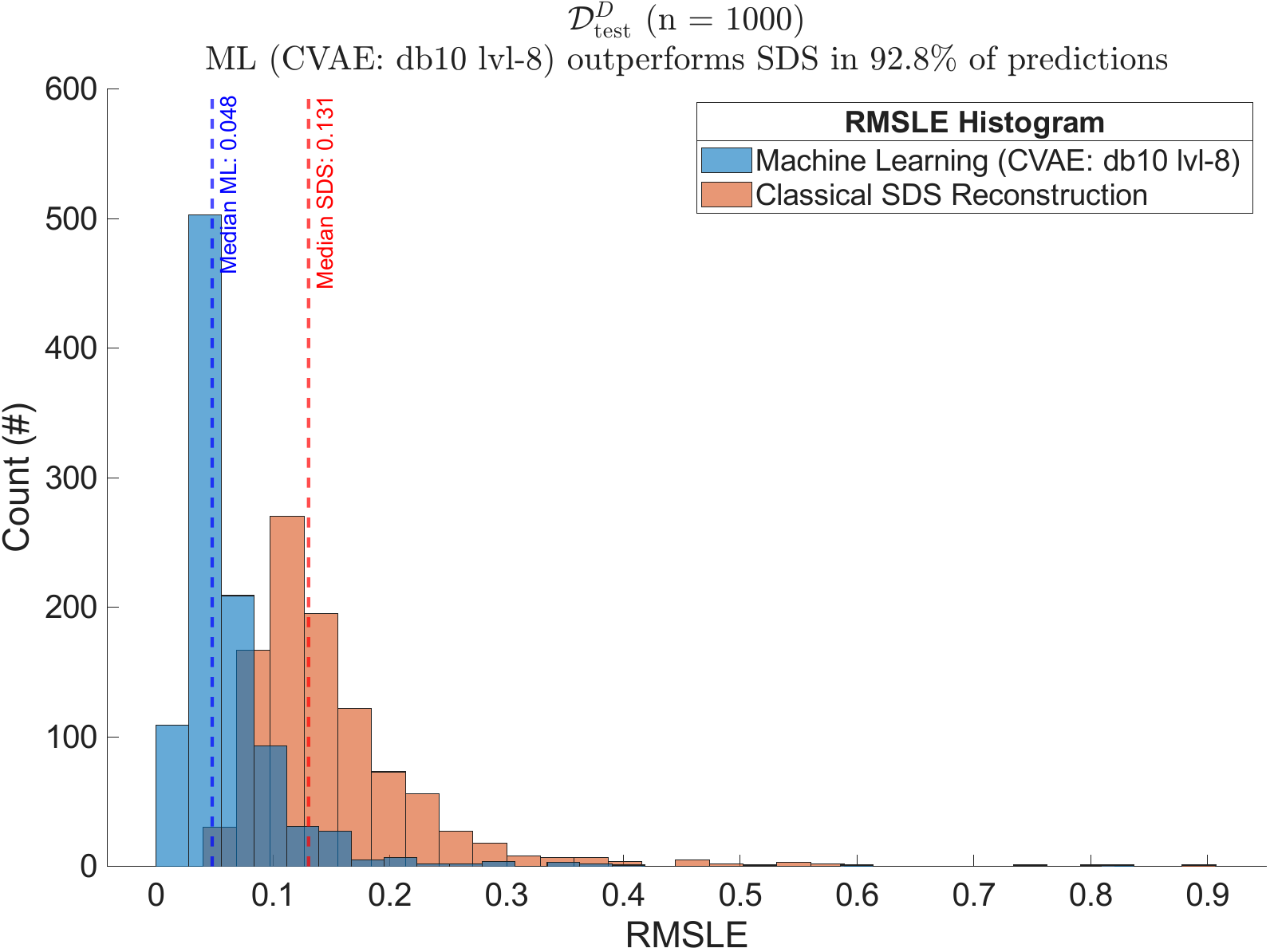}
    \end{subfigure}\par

    \caption{Distributions of per-sample RMSLE for a single CVAE ensemble
    member using the Daubechies~10 (\texttt{db10}) wavelet at decomposition
    level~8 (blue) and classical SDS reconstructions (orange).
    The four held-out datasets are
    $\mathcal{D}_{\mathrm{test}}^{A}$ ($n=963$, top left),
    $\mathcal{D}_{\mathrm{test}}^{B}$ ($n=890$, top right),
    $\mathcal{D}_{\mathrm{test}}^{C}$ ($n=815$, bottom left), and
    $\mathcal{D}_{\mathrm{test}}^{D}$ ($n=1000$, bottom right).
    The CVAE achieves lower RMSLE than SDS for 98.1\%, 98.9\%, 90.8\%,
    and 92.8\% of samples, respectively.
    Vertical dashed lines indicate the median RMSLE for each method.
    Lower RMSLE indicates closer agreement with the target SRS.}
    \label{fig:Holdout_RMSLE_ABCD}
\end{figure}

\begin{figure}[H]
    \centering
    \begin{subfigure}{0.48\textwidth}
        \centering
        \includegraphics[width=\linewidth]{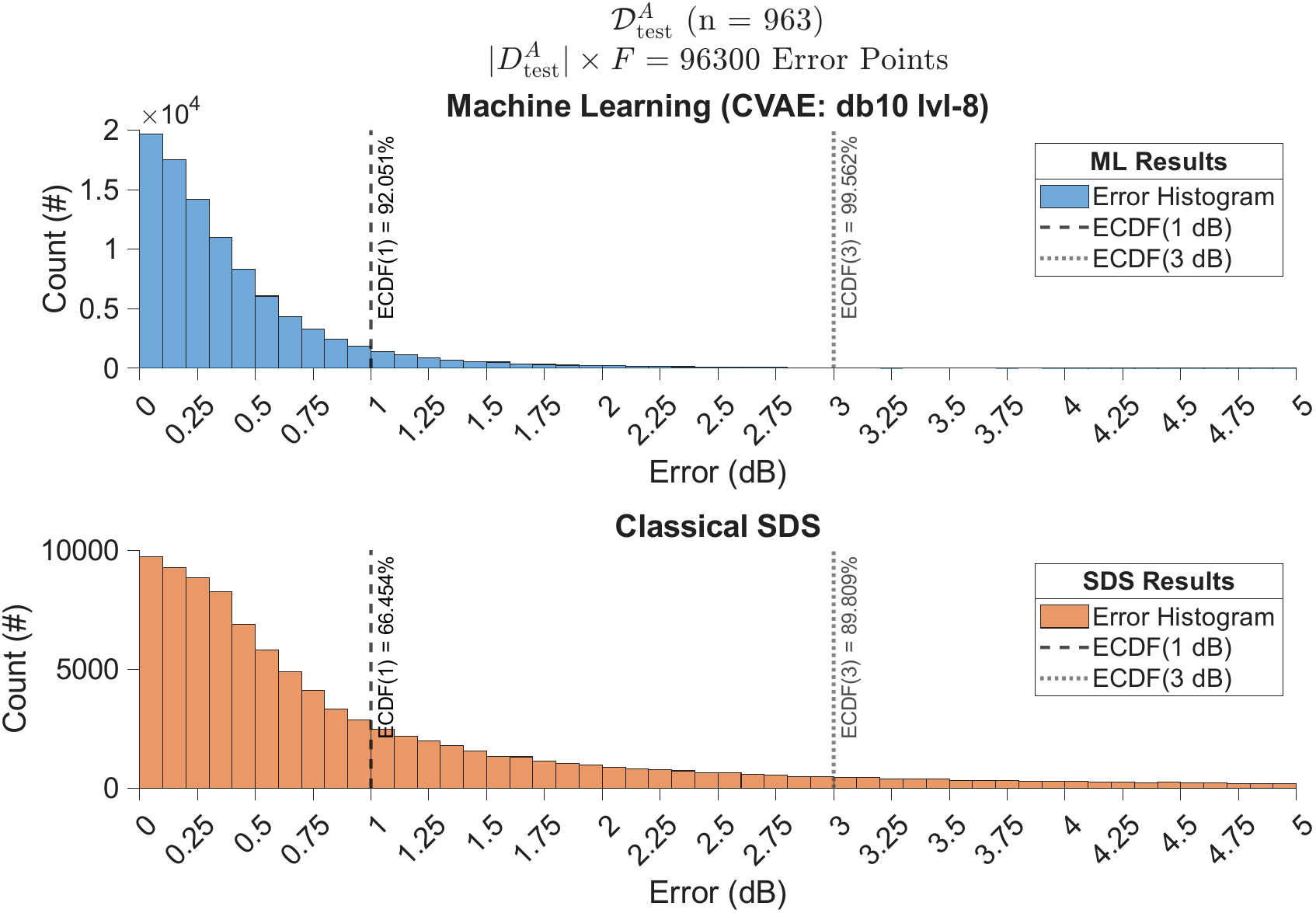}
    \end{subfigure}\hfill%
    \begin{subfigure}{0.48\textwidth}
        \centering
        \includegraphics[width=\linewidth]{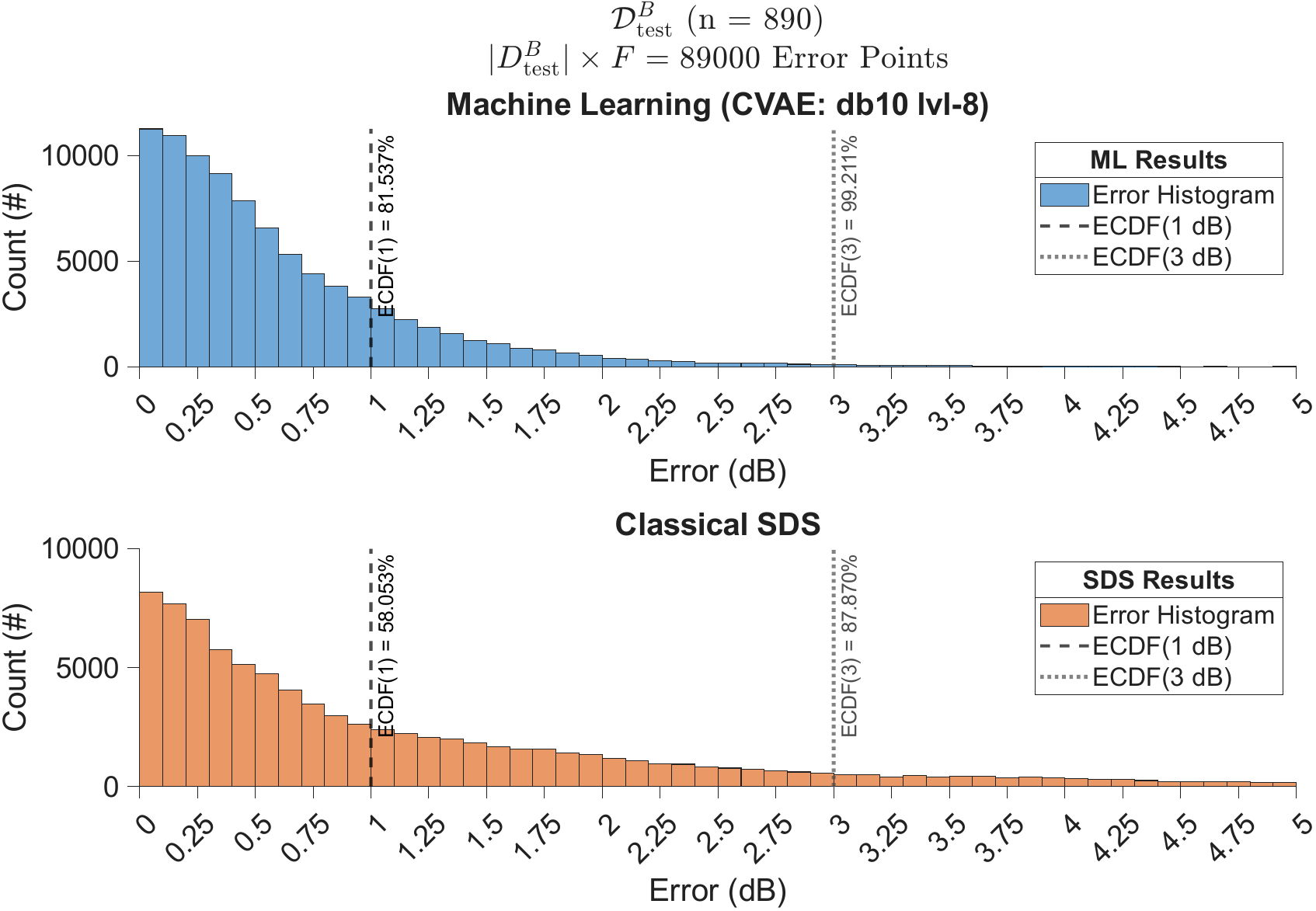}
    \end{subfigure}\par
    \vspace{1em}

    \begin{subfigure}{0.48\textwidth}
        \centering
        \includegraphics[width=\linewidth]{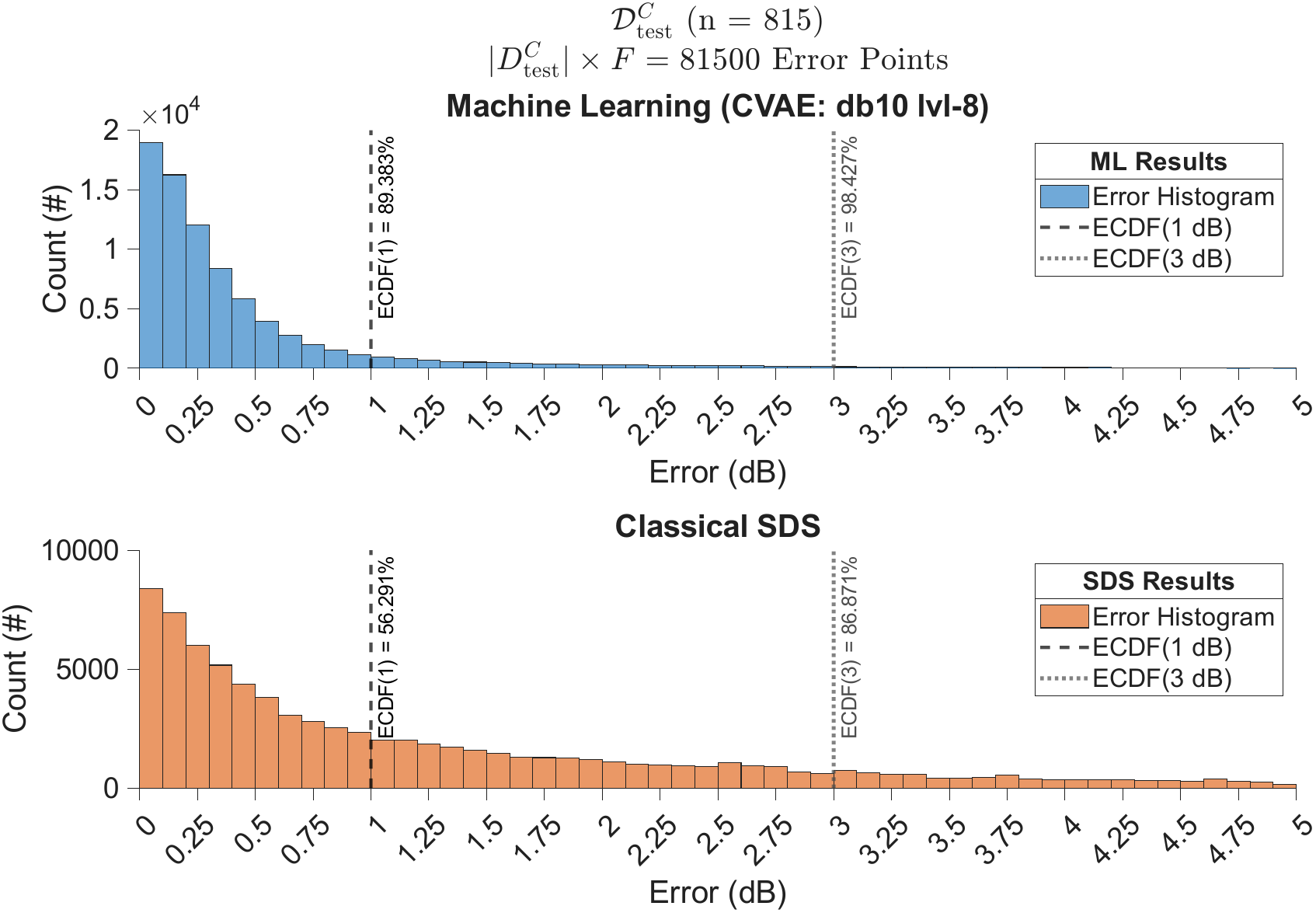}
    \end{subfigure}\hfill%
    \begin{subfigure}{0.48\textwidth}
        \centering
        \includegraphics[width=\linewidth]{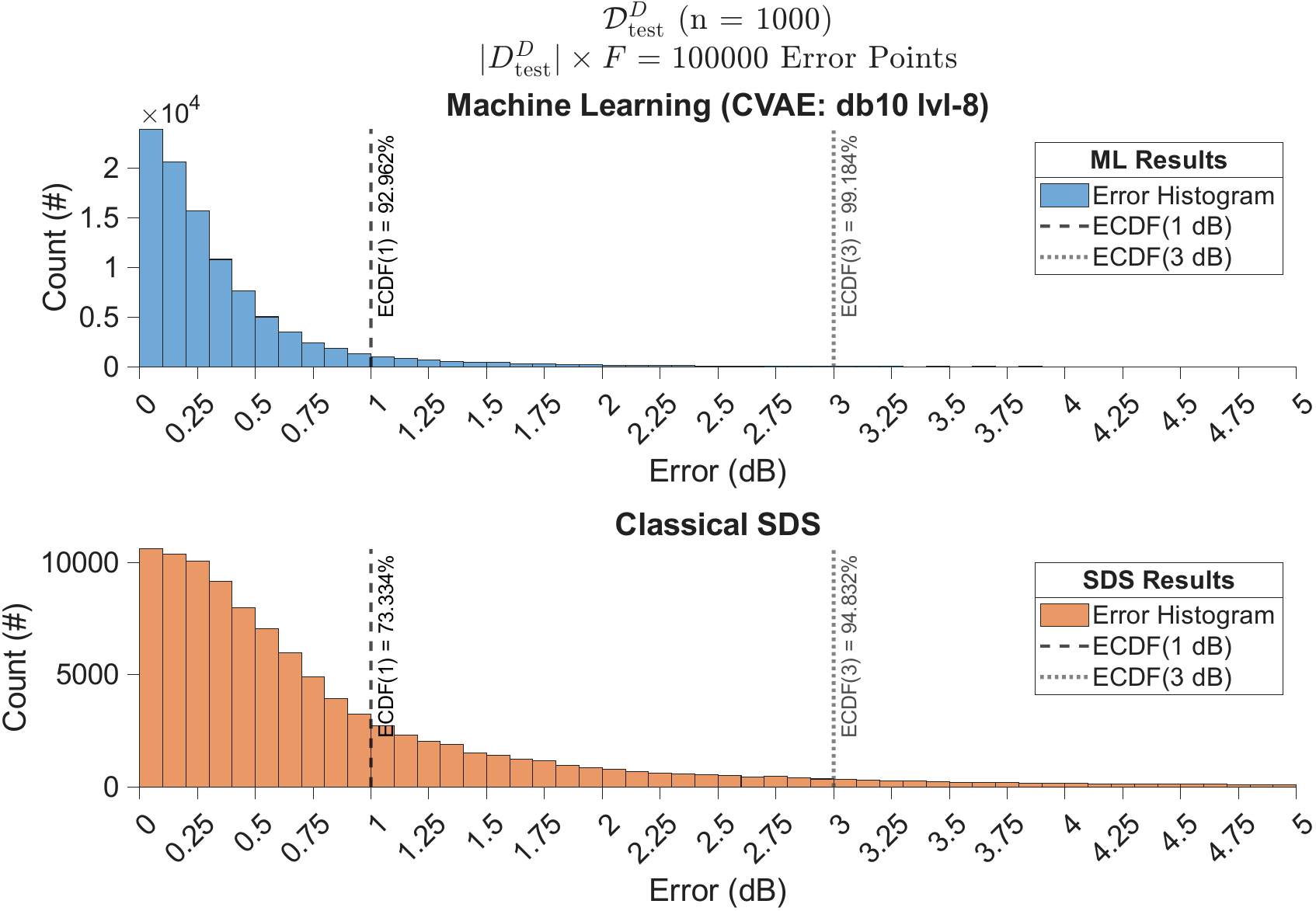}
    \end{subfigure}\par

    \caption{Distributions of per-frequency absolute SRS errors for a single
    CVAE ensemble member using the Daubechies~10 (\texttt{db10}) wavelet at
    decomposition level~8 (blue) and classical SDS reconstructions (orange).
    The four held-out datasets are
    $\mathcal{D}_{\mathrm{test}}^{A}$ (top left),
    $\mathcal{D}_{\mathrm{test}}^{B}$ (top right),
    $\mathcal{D}_{\mathrm{test}}^{C}$ (bottom left), and
    $\mathcal{D}_{\mathrm{test}}^{D}$ (bottom right).
    Errors are pooled across all samples and $F=100$ evaluation natural
    frequencies, yielding $|\mathcal{D}_{j}^{\mathrm{test}}|\times F$
    error values per method for each dataset.
    Vertical lines mark the 1~dB and 3~dB thresholds, with annotations
    reporting the corresponding empirical cumulative fractions.
    The CVAE achieves higher fractions of SRS points within both
    thresholds across all four datasets.}
    \label{fig:Holdout_dbEror_ABCD}
\end{figure}

The wavelet model reduced mean RMSLE by 31.6\%, 30.9\%, 23.0\%, and 34.7\% relative to the direct CVAE on datasets A--D. Medians and 97.5th percentiles also decreased, but maximum errors increased on A, C, and D. Thus, the gains did not extend uniformly to extreme errors. Figure~\ref{fig:Holdout_RMSLE_ABCD} shows per-sample RMSLE distributions, and Figure~\ref{fig:Holdout_dbEror_ABCD} shows pooled per-frequency errors for the \texttt{db10}, level~8 model. 

\ref{app:heldout_rmsle_results} summarizes median per-sample RMSLE across the tested wavelet families, orders, and decomposition levels. For many wavelets, median RMSLE decreased as decomposition level increased, although the trend was not monotonic and some configurations exhibited higher errors at the deepest tested levels. The best observed level varied with the wavelet and benchmark dataset, and higher wavelet order did not consistently yield lower error. All evaluated configurations achieved lower median RMSLE than SDS across the four benchmark datasets.

\ref{app:heldout_db_results} presents complementary results using the percentages of pooled SRS points within 1~dB and 3~dB of the target. These percentages generally improved with decomposition level, with variation across wavelets and datasets.

The number of tested configurations varied across wavelet families because computational constraints limited the model-training sweep. Each model required approximately one day of training on a single NVIDIA H200 or RTX PRO 6000 Blackwell Workstation Edition GPU, amounting to over 100 GPU-days across the model library.

To assess whether the CVAE provides an accuracy benefit despite its limited sensitivity to latent sampling, a preliminary comparison with deterministic wavelet regression used Haar (\texttt{db1}) at level~6 and otherwise identical training and evaluation conditions. The CVAE achieved lower mean and median RMSLE on all four datasets, reducing the mean by approximately 20--42\%. These results suggest that the CVAE formulation retains a reconstruction-accuracy benefit in this configuration even when latent sampling provides little waveform diversity.

Beyond direct comparisons of SRS accuracy, the learned approach changes how waveform synthesis is parameterized and performed. Conventional SDS methods optimize the parameters of a prescribed collection of decaying sinusoidal components for each target. The wavelet CVAE instead learns to predict coefficients across scales and temporal locations, allowing waveforms beyond this prescribed component family while retaining the structural assumptions of the wavelet decoder. Training draws on the entire model-development dataset, incorporating information from both synthetic and measured shocks into a shared synthesis mapping. Once trained, the same model and inference settings are applied across targets, without per-target parameter optimization or manual tuning.

Learned synthesis and conventional optimization are also complementary. Generated waveforms could serve as initial candidates for subsequent refinement to improve spectral agreement or satisfy additional application-specific constraints. This possibility motivates future work combining rapid learned synthesis with targeted optimization.

\subsection{SRS Conditioning Configuration Study}
\label{sec:srs_conditioning_study}

The conditioning study used the same \texttt{db10}, level~8 model configuration presented in \ref{sec:Representative} and \ref{sec:performance}, maintaining a consistent wavelet and decomposition level across these analyses. Synthetic training data used component frequencies sampled from 2–6000 Hz, as in the held-out model library. We compared three SRS conditioning grids: 100 log-random frequencies, 10 fixed frequencies, and a fixed 100-point grid adding 90 interpolated ordinates to those 10 points. The target spectrum was taken from Table~1 of Brake~\cite{brake_inverse_2011}; the conditioning-study results are reported in Table~\ref{tab:BrakeComparison} of the present study. Fixed grids used both concatenation and Hadamard encoding; the variable grid used concatenation. Other architectural and training settings were held constant.

For the log-random configuration, each training example was assigned 100 natural frequencies randomly sampled in logarithmic space and held fixed throughout training. At inference, all 100-point models used the same fixed conditioning grid and interpolated target ordinates. This grid retained the 10 published frequencies and added 90 points, allocated among adjacent intervals in proportion to their logarithmic frequency widths using the largest-remainder method. Additional points were logarithmically spaced within each interval. All target interpolation used SRS amplitudes linear in $\log_{10}$ frequency, preserving the published ordinates exactly.

The SRS conditioning grids spanned natural frequencies from 10~Hz to 5~kHz. This upper 5kHz limit was selected as a computational compromise, allowing SRS calculations at the native waveform sampling rate of 32{,}768~Hz without dynamic upsampling. At 5~kHz, this provides approximately 6.55 samples per oscillator cycle, below the eight-samples-per-cycle guideline \cite{lalanne_mechanical_2014}. The original target's 10~kHz and 20~kHz ordinates from Table~1 of Brake~\cite{brake_inverse_2011} were therefore excluded, and reconstruction errors were evaluated at the 10 remaining published natural frequencies. 

\begin{table*}[htbp]
\centering
\caption{Reconstructed SRS ordinates and corresponding error metrics
for the portion of the target spectrum from Table~1 of Brake
\cite{brake_inverse_2011} spanning 10~Hz to 5~kHz. The symbols
$\mathrm{C}$ and $\mathrm{H}$ denote concatenation and Hadamard
encoding, respectively. Errors are evaluated at the 10 retained
frequencies reported by Brake.}
\label{tab:BrakeComparison}

\resizebox{\textwidth}{!}{%
\begin{tabular}{lrrrrrrr}
\toprule
& & & \multicolumn{5}{c}{CVAE reconstruction} \\
\cmidrule(lr){4-8}
& & &
\multicolumn{1}{c}{Log-random 100-point grid} &
\multicolumn{2}{c}{Fixed 10-point grid} &
\multicolumn{2}{c}{Fixed 100-point grid} \\
\cmidrule(lr){4-4}
\cmidrule(lr){5-6}
\cmidrule(lr){7-8}
Frequency
& \multicolumn{1}{c}{Target}
& \multicolumn{1}{c}{Brake}
& \multicolumn{1}{c}{$\mathrm{C}$}
& \multicolumn{1}{c}{$\mathrm{C}$}
& \multicolumn{1}{c}{$\mathrm{H}$}
& \multicolumn{1}{c}{$\mathrm{C}$}
& \multicolumn{1}{c}{$\mathrm{H}$} \\
(Hz)
& 
& \multicolumn{1}{c}{(2011)}
& & & & & \\
\midrule

\multicolumn{8}{l}{
    \textit{Panel A: Reconstructed SRS ordinates (m/s$^2$)}
} \\
\addlinespace

10   & 2.000  & 1.960  & 1.983  & 2.057  & 1.958  & 1.844  & 1.956 \\
100  & 6.000  & 5.709  & 5.825  & 5.958  & 5.793  & 5.781  & 5.741 \\
400  & 6.000  & 5.937  & 6.126  & 5.990  & 6.049  & 6.250  & 6.187 \\
500  & 9.000  & 9.350  & 8.644  & 9.060  & 9.418  & 8.903  & 8.401 \\
600  & 9.000  & 7.484  & 8.369  & 9.357  & 8.790  & 8.528  & 8.588 \\
700  & 6.000  & 6.309  & 6.856  & 5.852  & 6.226  & 6.427  & 6.736 \\
2000 & 8.000  & 8.044  & 8.788  & 8.019  & 8.063  & 8.183  & 7.657 \\
2500 & 11.000 & 10.270 & 10.589 & 11.017 & 10.940 & 10.856 & 10.298 \\
3000 & 8.000  & 8.289  & 8.744  & 8.060  & 8.118  & 8.334  & 8.329 \\
5000 & 8.000  & 8.400  & 7.275  & 7.861  & 8.297  & 8.242  & 8.809 \\

\midrule
\multicolumn{8}{l}{
    \textit{Panel B: Error relative to the target SRS (dB)}
} \\
\addlinespace

MAE
& -- & 0.444 & 0.540 & \textbf{0.118} & 0.204 & 0.347 & 0.499 \\

RMSE
& -- & 0.610 & 0.634 & \textbf{0.160} & 0.236 & 0.394 & 0.555 \\

Maximum absolute error
& -- & 1.602 & 1.159 & \textbf{0.338} & 0.394 & 0.704 & 1.004 \\

\bottomrule
\end{tabular}%
}
\end{table*}

The fixed 10-point grid with concatenation achieved the lowest MAE, RMSE, and maximum absolute error. Concatenation also outperformed Hadamard encoding on the fixed 100-point grid. These results motivated concatenation in the subsequent comparisons. These findings suggest a tradeoff between reconstruction accuracy on fixed grids and generalizability across target natural-frequency grids, as evaluated using the log-random 100-point configuration. The 10-point result applies at the evaluated ordinates: the 100-point models additionally represented interpolated targets, and accuracy between the published frequencies was not established.

\subsection{Comparison with Brake (2011)}

Retaining the \texttt{db10}, level~8 configuration, three independently trained CVAEs with concatenation conditioning were evaluated against the results reported by Brake \cite{brake_inverse_2011} for four tabulated target spectra. Because these targets extend to natural frequencies of 10~kHz, the synthetic component frequency range was changed from 2--6000~Hz to 10--15{,}000~Hz to provide higher-frequency content in the training waveforms. The measured training data and synthetic generation procedure were otherwise unchanged from those used for the benchmark models. One model used the retained grid from Brake's Table~1, another used Table~2, and the third used the common grid of Tables~3 and~4, matching training constraints to the evaluated ordinates.

Generated waveforms were sampled at 32{,}768~Hz and upsampled using Fourier interpolation before SRS evaluation. The integer upsampling factor was selected to provide at least eight samples per cycle at the highest evaluated natural frequency \cite{lalanne_mechanical_2014}. For the 10~kHz upper evaluation frequency, this yielded a factor of three and an effective sampling rate of 98{,}304~Hz. Interpolation adds no signal content above the native Nyquist frequency of 16{,}384~Hz \cite{schafer_digital_1973}. We conservatively excluded each target's 20~kHz ordinate and evaluated all methods at the same retained frequencies within 10~Hz--10~kHz. \ref{app:brake-comparisons} reports the ordinates and errors.

The CVAE achieved lower MAE, RMSE, and maximum absolute error than Brake's non-GA reconstructions for three of four targets; the exception was Table~2, with five retained frequencies. Including the GA-refined results available for Tables~3 and~4, the CVAE had the lowest values of all three metrics for Tables~1 and~3, but the GA-refined reconstruction performed better for Table~4. The advantage therefore depended on the target. These comparisons assess spectral agreement; the distinct ensemble waveforms are illustrated separately in Figure \ref{fig:representative-ensemble}.

Brake's tabulated targets span narrower SRS amplitude ranges than the representative benchmark spectrum shown in Figure~\ref{fig:representative-ensemble}. These comparisons therefore complement the held-out benchmarks by assessing agreement with published targets, while the benchmarks evaluate performance across a broader collection of measured and synthetic spectral shapes.

\subsection{Computational Performance}

Single-model inference on an NVIDIA RTX 6000 Ada Generation workstation GPU, using a batch size of 64, required an average of 0.1--0.6~ms per waveform. Batch size was not optimized for throughput. SDS required 5--8~s per target on an AMD Ryzen~9 9900X 12-core CPU. These timings reflect the respective execution configurations. An internal SDS+GA implementation required 20--30~min per target, making evaluation over thousands of benchmark spectra impractical. Its timing is provided for context; benchmark accuracy comparisons use SDS without GA. Brake's implementation was unavailable for direct timing comparisons.

These inference timings exclude training. On an NVIDIA H200 GPU, training required approximately 8~h for models with 5--8 conditioning frequencies and 24~h for models with 100 frequencies. Denser grids or higher target frequencies can increase this cost. Trained models can be reused within their supported configurations, making generation from multiple ensemble members practical.

\section{Conclusion}

A recurrent wavelet decoder extends our earlier inverse-SRS CVAE by predicting multiscale wavelet coefficients rather than predicting  time-domain samples directly. The evaluated \texttt{db10}, level~8 model reduced mean RMSLE by 23--35\% relative to the earlier CVAE across four held-out datasets and outperformed SDS in mean and median error. Extreme errors did not improve uniformly, and comparisons with Brake's reconstructions favored the CVAE for some targets but not all. A preliminary deterministic-regression comparison also favored the CVAE.

Concatenation conditioning improved accuracy on the evaluated fixed grids; whether these gains extend across other wavelet configurations remains to be established. Inference required 0.1--0.6~ms per waveform versus 5--8~s for SDS, excluding offline training.

Limited variability under latent sampling motivated independently trained wavelet models that produce distinct candidates for a common target. Spectral agreement alone does not establish physical suitability or coverage of the inverse solution space. Future work will quantify diversity, temporal localization, duration, energy, amplitude statistics, and frequency content against application-specific constraints, and investigate newer generative models for synthesizing diverse candidate waveforms within a single model. A longer-term goal is to develop a foundation model for shock time-series synthesis, trained on a broader corpus of measured and synthetic transients and adaptable to different shock environments, conditioning information, and test requirements.

\section*{CRediT authorship contribution statement}

\textbf{Adam Watts:} Conceptualization, Methodology, Software,
Validation, Formal analysis, Investigation, Resources, Data curation,
Writing -- original draft, Writing -- review \& editing, Visualization,
Supervision, Project administration, Funding acquisition.
\textbf{Charles Young:} Conceptualization, Supervision,
Writing -- review \& editing.
\textbf{Ryan Bowering:} Conceptualization, Methodology, Software,
Investigation.
\textbf{Andrew Jeon:} Conceptualization, Investigation,
Writing -- review \& editing.
\textbf{Destry Newton:} Conceptualization, Investigation,
Writing -- review \& editing.

\section*{Declaration of competing interest}
The authors declare that they have no known competing financial interests or personal relationships that could have appeared
to influence the work reported in this paper.

\section*{Acknowledgments}
This work was supported by the U.S. Department of Energy through the Los Alamos National Laboratory. Los Alamos National Laboratory is operated by Triad National Security, LLC, for the National Nuclear Security Administration of U.S. Department of Energy (Contract No. 89233218CNA000001).

\appendix

\section{Training Objective and Loss Implementation}
\label{app:loss_implementation}

This appendix defines the components of the training objective in Eq.~\eqref{eq:ltotal}. The time-series, PSD, SRS, and KL divergence terms follow our earlier framework~\cite{watts2026inverse}.  The accelerated waveform-shape implementation used in both studies, but not detailed in the earlier paper, is described below.

\subsection{Waveform-Shape Loss}
\label{app:waveform_shape}

The waveform-shape loss compares absolute SDOF acceleration response histories near their respective maxima. Responses are evaluated at the natural frequencies specified for each model configuration.

Let $a_i[n]=|r_i[n]|$ and $\hat{a}_i[n]=|\hat{r}_i[n]|$ denote the reference and generated response magnitudes at natural frequency $\mathfrak{f}_i$. Local windows $A_i[k]$ and $\hat{A}_i[k]$ are gathered around their respective maximum-magnitude sample indices, $p_i$ and $\hat{p}_i$, so that $k=0$ aligns the two maxima. The first maximum is selected when peaks are tied, and out-of-range sample indices are clipped to the nearest record endpoint.

Soft peak locations are additionally computed as
\[
\mu_i
=
\sum_{n=0}^{T-1}
n\,\operatorname{softmax}_{n}\!\left(a_i[n]/\tau\right),
\]
with $\hat{\mu}_i$ defined analogously for the generated response. Here, $T$ includes residual-response padding and $\tau=0.03$. These locations inform window sizing and weighting; the sampled windows remain aligned at the integer maxima.

The nominal Gaussian width is $\sigma_{i,\mathrm{raw}}=(c/6)(f_s/\mathfrak{f}_i)$ samples, with $c=1.5$ and $f_s=32768$~Hz. The implementation uses a nominal half-width of $K\sigma_i$, with $K=2.5$, a nominal minimum Gaussian width of two samples, and approximate window-length bounds of 1--10\% of $T$. The width is constrained by these bounds, and the window is shortened where required by the soft peak locations' proximity to the record boundaries.

For sample offsets $k$ in the resulting window $\mathcal{W}_i$, the Gaussian weights are
\[
g_i[k]
=
\exp\!\left[-\frac{(k+p_i-\mu_i)^2}{2\sigma_i^2}\right],
\qquad
W_i[k]
=
\frac{g_i[k]}
{\sum_{\ell\in\mathcal{W}_i}g_i[\ell]+10^{-12}}.
\]
Thus, the Gaussian is centered using the reference soft peak location, and the same weights are applied to both aligned windows. A one-sample window uses a point weight at $k=0$.

The per-frequency discrepancy and scalar loss are
\begin{equation}
\mathcal{E}_{\mathrm{shape},i}
=
\sum_{k\in\mathcal{W}_i}
W_i[k]\left(\hat{A}_i[k]-A_i[k]\right)^2,
\label{eq:srs_shape}
\end{equation}
\begin{equation}
\mathcal{L}_{\mathrm{shape}}
=
\frac{1}{F}\sum_{i=1}^{F}
\mathcal{E}_{\mathrm{shape},i}.
\label{eq:lshape}
\end{equation}
The implementation sums over window samples and averages over frequencies and the batch; no additional division by window length or logarithmic transformation is applied.

For one-sample windows, the loss becomes, up to numerical stabilization, mean squared error between maximax SRS ordinates. Wider windows also penalize differences in the surrounding response magnitudes. This differs from the logarithmic SRS loss. Soft weighting does not remove the discrete operations used for integer peak selection and window sizing.

\subsection{Time-Series, PSD, and SRS Losses}

The time-series loss compares the signed reference and generated
acceleration samples using root-mean-squared error:
\begin{equation}
\mathcal{L}_{\mathrm{TS}}
=
\operatorname{RMSE}(\mathbf{x},\hat{\mathbf{x}})
=
\sqrt{
\frac{1}{N}\sum_{n=0}^{N-1}
\left(\hat{x}[n]-x[n]\right)^2
}.
\label{eq:lts}
\end{equation}
Here, $N$ is the shock time-series length, distinct from the padded SDOF response-record length $T$.

For nonnegative vectors
$\mathbf{y},\hat{\mathbf{y}}\in\mathbb{R}_{\geq0}^{M}$,
the mean squared logarithmic error is
\begin{equation}
\operatorname{MSLE}(\mathbf{y},\hat{\mathbf{y}})
=
\frac{1}{M}\sum_{j=1}^{M}
\left[
\log(\hat{y}_j+\epsilon)-\log(y_j+\epsilon)
\right]^2,
\label{eq:msle}
\end{equation}
where $\epsilon>0$ provides numerical stability.

The PSD loss compares the Welch PSD estimates
$\mathbf{P}_{xx}$ and $\mathbf{P}_{\hat{x}\hat{x}}$:
\begin{equation}
\mathcal{L}_{\mathrm{PSD}}
=
\operatorname{MSLE}
\left(\mathbf{P}_{xx},\mathbf{P}_{\hat{x}\hat{x}}\right).
\label{eq:lpsd}
\end{equation}

The SRS loss is
\begin{equation}
\mathcal{L}_{\mathrm{SRS}}
=
\operatorname{MSLE}
\left(
\operatorname{SRS}\{\mathbf{x},\boldsymbol{\mathfrak{f}}\},
\operatorname{SRS}\{\hat{\mathbf{x}},\boldsymbol{\mathfrak{f}}\}
\right),
\label{eq:lsrs}
\end{equation}
where $\boldsymbol{\mathfrak{f}}=(\mathfrak{f}_1,\ldots,\mathfrak{f}_F)$ contains the evaluation natural frequencies.

\subsection{KL Divergence}

For a diagonal Gaussian approximate posterior and a standard normal prior, the KL divergence is
\begin{equation}
\begin{aligned}
\mathcal{L}_{\mathrm{KL}}
&=
D_{\mathrm{KL}}
\left[
q_{\phi}(\mathbf{z}\mid\mathbf{x},\mathbf{s})
\,\Vert\,\mathcal{N}(\mathbf{0},\mathbf{I})
\right] \\
&=
\frac{1}{2}\sum_{j=1}^{d_z}
\left(
\mu_{q,j}^{2}
+
\sigma_{q,j}^{2}
-
\log\sigma_{q,j}^{2}
-
1
\right).
\end{aligned}
\label{EQ:Simple_KL}
\end{equation}
Here, $\mathbf{s}$ denotes the SRS conditioning information, $d_z$ is the latent-space dimension, and $\mu_{q,j}$ and $\sigma_{q,j}^{2}$ are the posterior mean and variance for latent component $j$.

\subsection{Loss Coefficients}

The coefficients in Eq.~\eqref{eq:ltotal} were selected
empirically and are listed in Table~\ref{tab:loss_coefficients}.

\begin{table}[htbp]
\centering
\caption{Coefficients used in the total training objective.}
\label{tab:loss_coefficients}
\begin{tabular}{llr}
\hline
Loss term & Coefficient & Value \\
\hline
Waveform shape & $\lambda_{\mathrm{shape}}$ & 3.4601 \\
Time series & $\lambda_{\mathrm{TS}}$ & 0.7658 \\
Welch PSD & $\lambda_{\mathrm{PSD}}$ & 0.1885 \\
SRS & $\lambda_{\mathrm{SRS}}$ & 2.9037 \\
KL divergence & $\lambda_{\mathrm{KL}}$ & 4.955 \\
\hline
\end{tabular}
\end{table}

\FloatBarrier
\section{Synthetic Data-Generation Details}
\label{sec:synthetic_generation_appendix}

Each synthetic shock contains a random number of basis components,
\begin{equation}
\mathcal{B}\sim\operatorname{Unif}\{1,\ldots,10\}.
\end{equation}
For each component $i\in\{1,\ldots,\mathcal{B}\}$, the basis family $k_i\in\{1,2\}$ is sampled uniformly, where $k_i=1$ denotes an exponentially decaying sinusoid and $k_i=2$ denotes a modified Morlet-like pulse.

Let $T_s=1/f_s$ denote the sampling interval and let $m\in\{0,\ldots,N-1\}$ denote the local sample index of a basis component. The two basis functions are defined as
\begin{equation}
\label{EQ:BasisCases}
\psi_{i,k_i}[m]
=
\begin{cases}
A_i
\exp\left(-\lambda_i mT_s\right)
\sin\left(2\pi f_i mT_s+\phi_i\right),
& k_i=1, \\[1.2em]
A_i
\exp\left[
\eta_i\omega_i
\left(\ln(1+mT_s)-mT_s\right)
\right]
\cos\left(\omega_i mT_s+\phi_i\right),
& k_i=2,
\end{cases}
\end{equation}
where $\omega_i=2\pi f_i$. The amplitude $A_i$, frequency $f_i$, decay parameter $\lambda_i$, phase $\phi_i$, and shape parameter $\eta_i$ are randomly sampled according to the distributions reported in
Table~\ref{tab:DataGeneratorParams}.

The first component is assigned a randomly sampled padding fraction,
\begin{equation}
\xi_1\sim\mathcal{U}(0,0.75).
\end{equation}
For each subsequent component, the padding fraction is either inherited from the preceding component or sampled independently:
\begin{equation}
\label{EQ:PaddingFraction}
\xi_i
=
\begin{cases}
\xi_{i-1},
& \text{with probability }0.5,\\
\xi_i',\quad \xi_i'\sim\mathcal{U}(0,0.75),
& \text{with probability }0.5.
\end{cases}
\end{equation}
The corresponding starting index is
\begin{equation}
\label{EQ:BasisStartIndex}
n_{0,i}=\left\lceil \xi_i N\right\rceil.
\end{equation}
Thus, each component begins between the start of the time series and 75\% of its total length. Inheriting the preceding offset produces temporally clustered components, whereas independently sampling a new offset distributes components throughout the time series.

The complete synthetic acceleration time series is
\begin{equation}
\label{EQ:SyntheticData}
\ddot{x}_{\mathrm{synth}}[n]
=
\epsilon[n]
+
\sum_{i=1}^{\mathcal{B}}
\mathbb{I}\!\left(n\geq n_{0,i}\right)
\psi_{i,k_i}\!\left[n-n_{0,i}\right],
\qquad
n=0,\ldots,N-1.
\end{equation}
Here, $\mathbb{I}(\cdot)$ denotes the indicator function, which equals one when its argument is true and zero otherwise. The $i$-th component therefore contributes nothing before its starting index $n_{0,i}$ and becomes active when $n\geq n_{0,i}$. The shifted argument $n-n_{0,i}$ defines the evolution of each basis function relative to its assigned starting location. The term $\epsilon[n]$ represents additive zero-mean Gaussian noise:
\begin{equation}
\epsilon[n]\sim\mathcal{N}\!\left(0,\sigma_{\epsilon}^{2}\right).
\end{equation}

\begin{table}[htbp]
\centering
\small
\begin{tabularx}{\textwidth}{@{}lXl@{}}
\toprule
Symbol & Parameter & Value or sampling distribution \\
\midrule
$\mathcal{B}$ & Number of basis components
    & $\operatorname{Unif}\{1,\ldots,10\}$ \\
$\xi_i'$ & Independently sampled offset fraction
    & $\mathcal{U}(0,0.75)$ \\
$\gamma$ & Probability of inheriting the preceding offset
    & $0.5$ \\
$\phi_i$ & Phase (rad)
    & $\mathcal{U}(0,2\pi)$ \\
$f_i$ & Component frequency (Hz)
    & $\mathcal{U}(2,6000)$ \\
$\lambda_i$ & Sinusoidal decay constant ($\mathrm{s}^{-1}$)
    & $\mathcal{U}(0.004\pi f_i,0.2\pi f_i)$ \\
$\eta_i$ & Modified Morlet-like pulse shape parameter
    & $\mathcal{U}(0.01,10)$ \\
$\epsilon[n]$ & Background noise
    & $\mathcal{N}(0,\sigma_{\epsilon}^{2})$ \\
$A_i$ & Component amplitude ($g$)
    & $\mathcal{U}(0.25,10)$ \\
$\sigma_{\epsilon}^{2}$ & Noise variance ($g^2$), sampled once per time series
    & $\mathcal{U}(0.005,0.05)$ \\
\bottomrule
\end{tabularx}
\caption{Synthetic shock generation parameters.
$\mathcal{U}$ denotes a continuous uniform distribution and
$\operatorname{Unif}$ a discrete uniform distribution.
The component-frequency range is 2--6000~Hz unless otherwise
specified. Basis definitions and temporal placement rules are
given in the accompanying equations. Here, $g$ denotes standard gravitational acceleration.}
\label{tab:DataGeneratorParams}
\end{table}

\FloatBarrier
\section{Held-Out Median RMSLE by Wavelet Configuration}
\label{app:heldout_rmsle_results}

\begin{figure}[H]
    \centering
    \includegraphics[width=\linewidth]{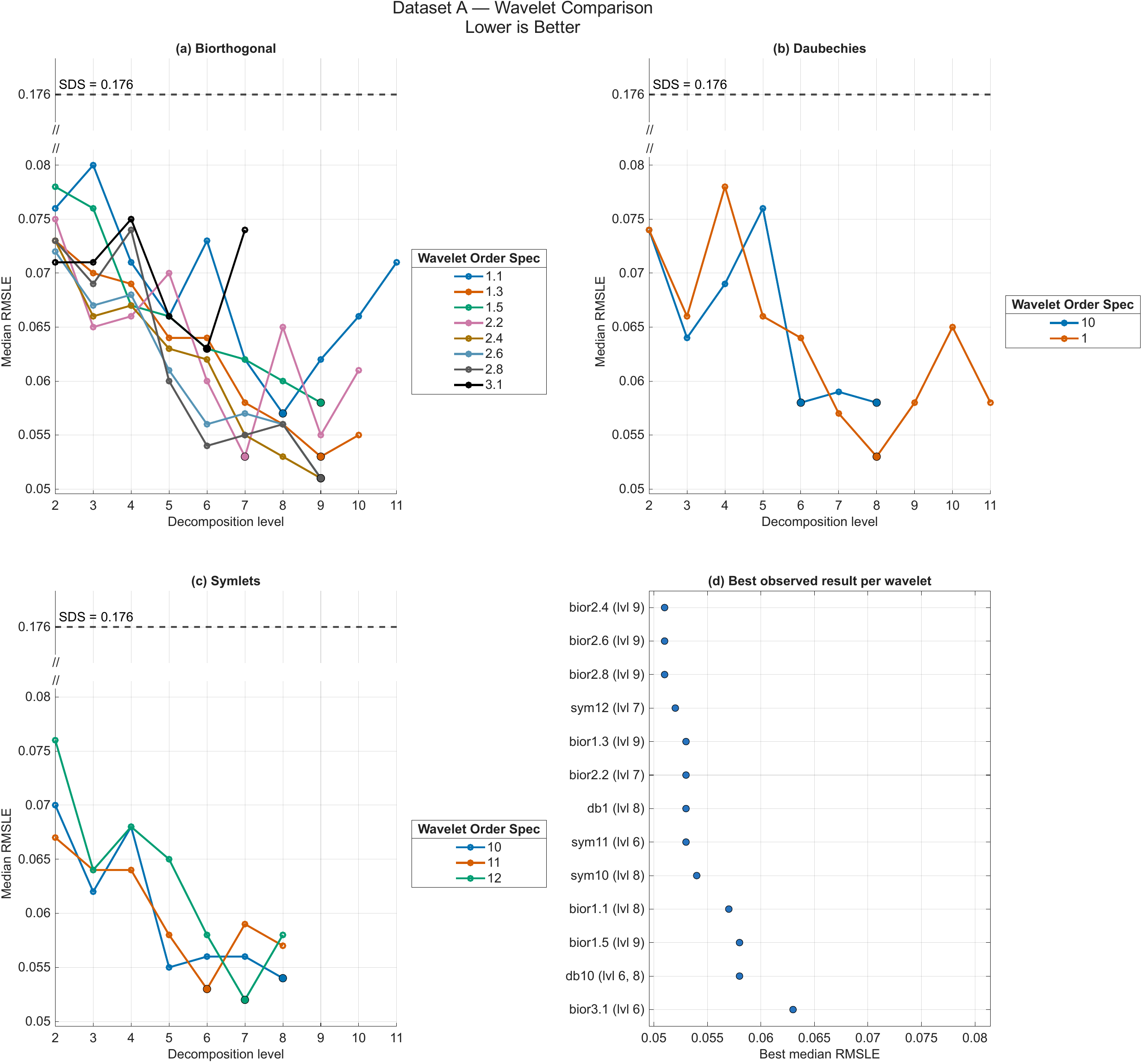}
    \caption{Held-out median RMSLE by wavelet configuration for Dataset A.}
    \label{fig:wavelet-comparison-a-rmsle}
\end{figure}

\begin{figure}[H]
    \centering
    \includegraphics[width=\linewidth]{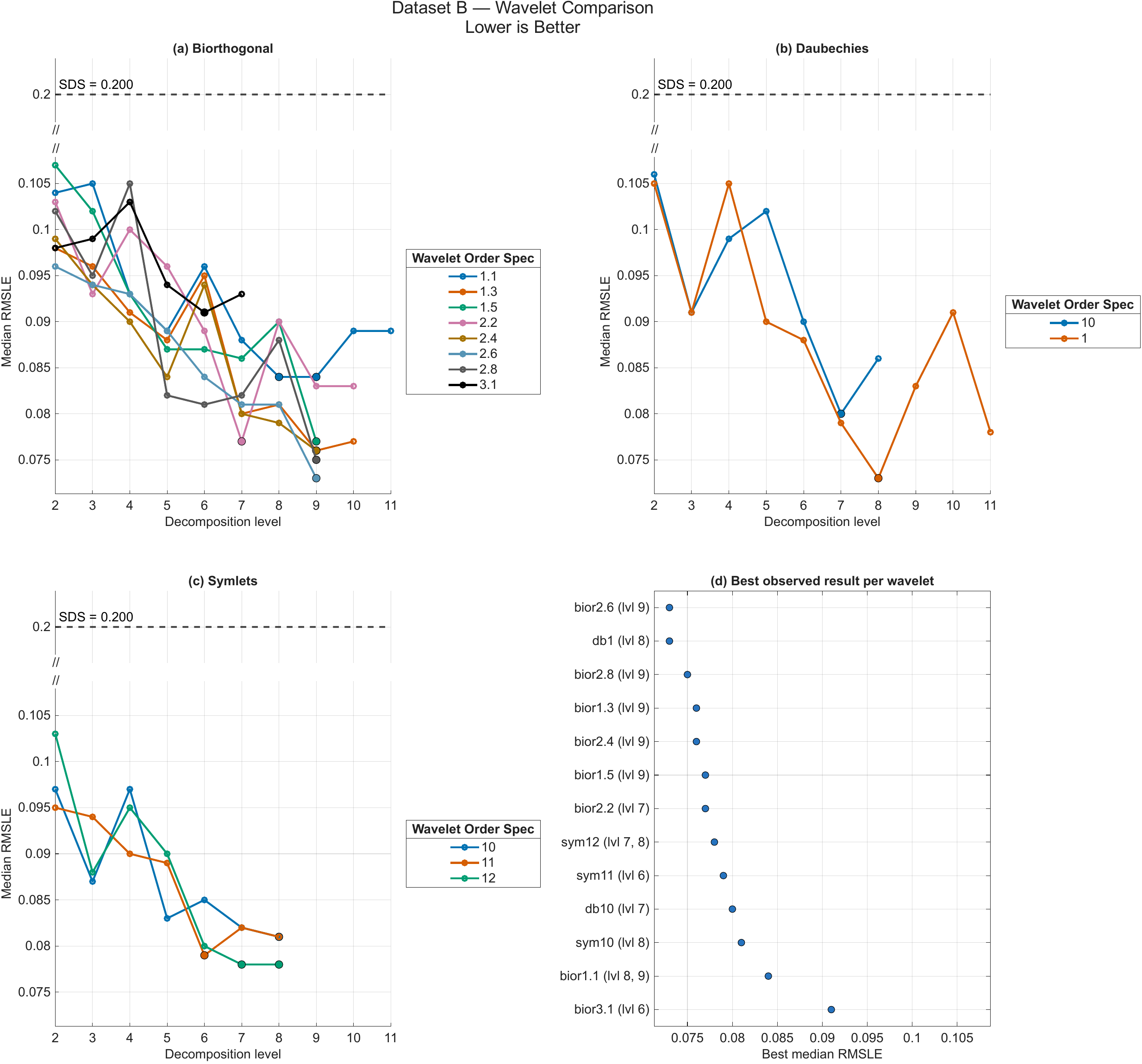}
    \caption{Held-out median RMSLE by wavelet configuration for Dataset B.}
    \label{fig:wavelet-comparison-b-rmsle}
\end{figure}

\begin{figure}[H]
    \centering
    \includegraphics[width=\linewidth]{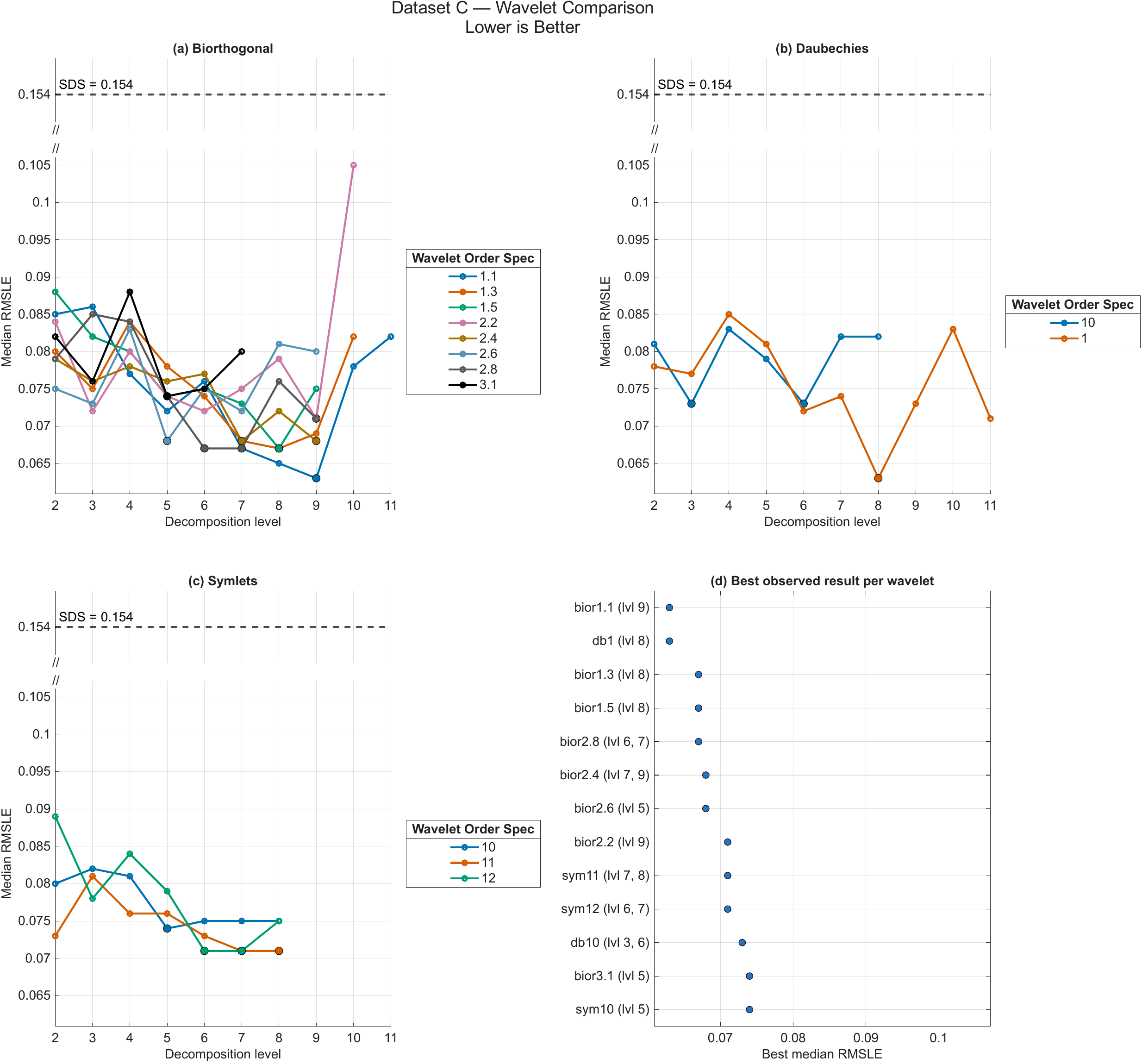}
    \caption{Held-out median RMSLE by wavelet configuration for Dataset C.}
    \label{fig:wavelet-comparison-c-rmsle}
\end{figure}

\begin{figure}[H]
    \centering
    \includegraphics[width=\linewidth]{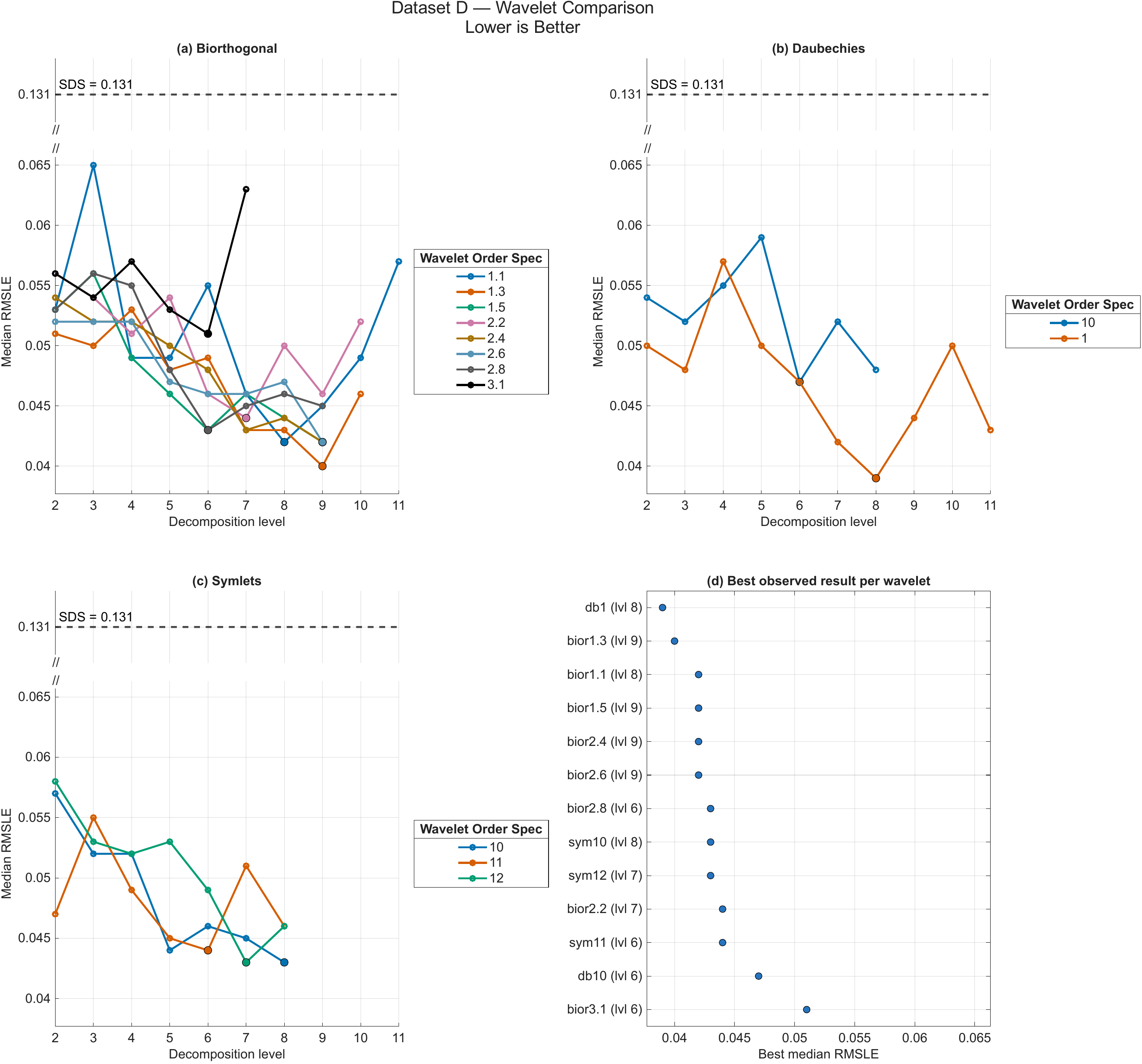}
    \caption{Held-out median RMSLE by wavelet configuration for Dataset D.}
    \label{fig:wavelet-comparison-d-rmsle}
\end{figure}

Figures~\ref{fig:wavelet-comparison-a-rmsle}--\ref{fig:wavelet-comparison-d-rmsle} summarize held-out median RMSLE across the tested wavelet configurations for Datasets A--D, respectively. In each figure, panel~(a) shows biorthogonal wavelets, panel~(b) Daubechies wavelets, and panel~(c) Symlet wavelets, with median RMSLE plotted against decomposition level. Each dataset's median CVAE RMSLE in Figure~\ref{fig:Holdout_RMSLE_ABCD} corresponds to the \texttt{db10}, level~8 point in panel~(b) of the corresponding dataset's wavelet-comparison figure. For example, the Dataset~A median of 0.058 appears at level~8 on the \texttt{db10} curve in Figure~\ref{fig:wavelet-comparison-a-rmsle}(b).

For the fixed 9000-sample signal length, maximum candidate decomposition levels vary with each wavelet's filter length, as described in
\ref{sec:Multi-Model Wavelet Ensemble}. Not all eligible configurations were trained because of computational cost; the plotted points represent configurations that were trained and evaluated. Filled markers indicate the lowest observed median RMSLE for each wavelet, including ties. Broken axes separate the SDS baseline from the wavelet scores. Panel (d) ranks each wavelet by its lowest observed median RMSLE across the tested levels. Lower RMSLE indicates better performance. 

\section{Held-Out dB Error by Wavelet Configuration}
\label{app:heldout_db_results}

\begin{figure}[H]
    \centering
    \includegraphics[width=\linewidth]{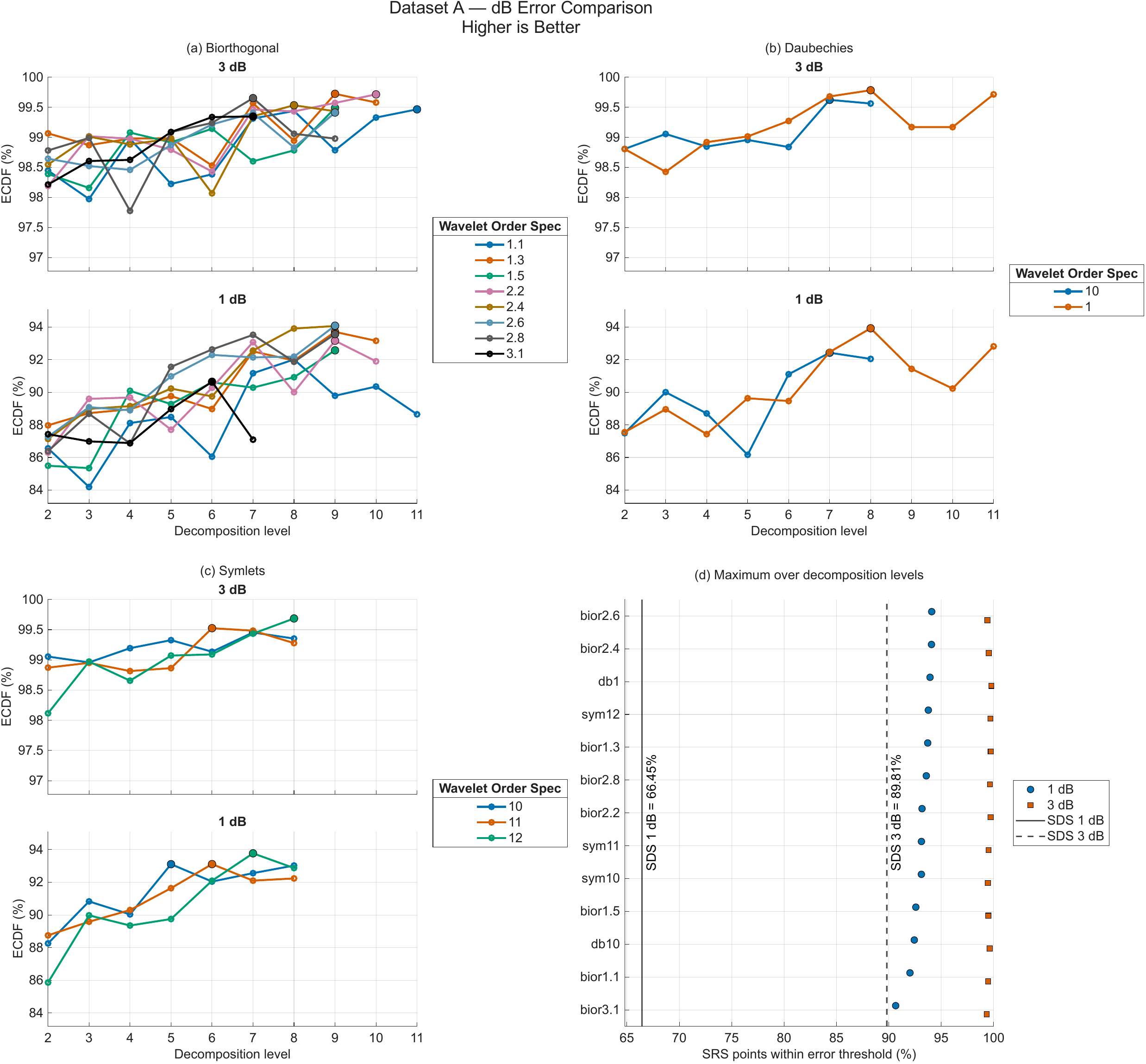}
    \caption{Percentage of pooled SRS points within 1~dB and 3~dB error thresholds by wavelet configuration for Dataset A.}
    \label{fig:wavelet-comparison-a-dB}
\end{figure}

\begin{figure}[H]
    \centering
    \includegraphics[width=\linewidth]{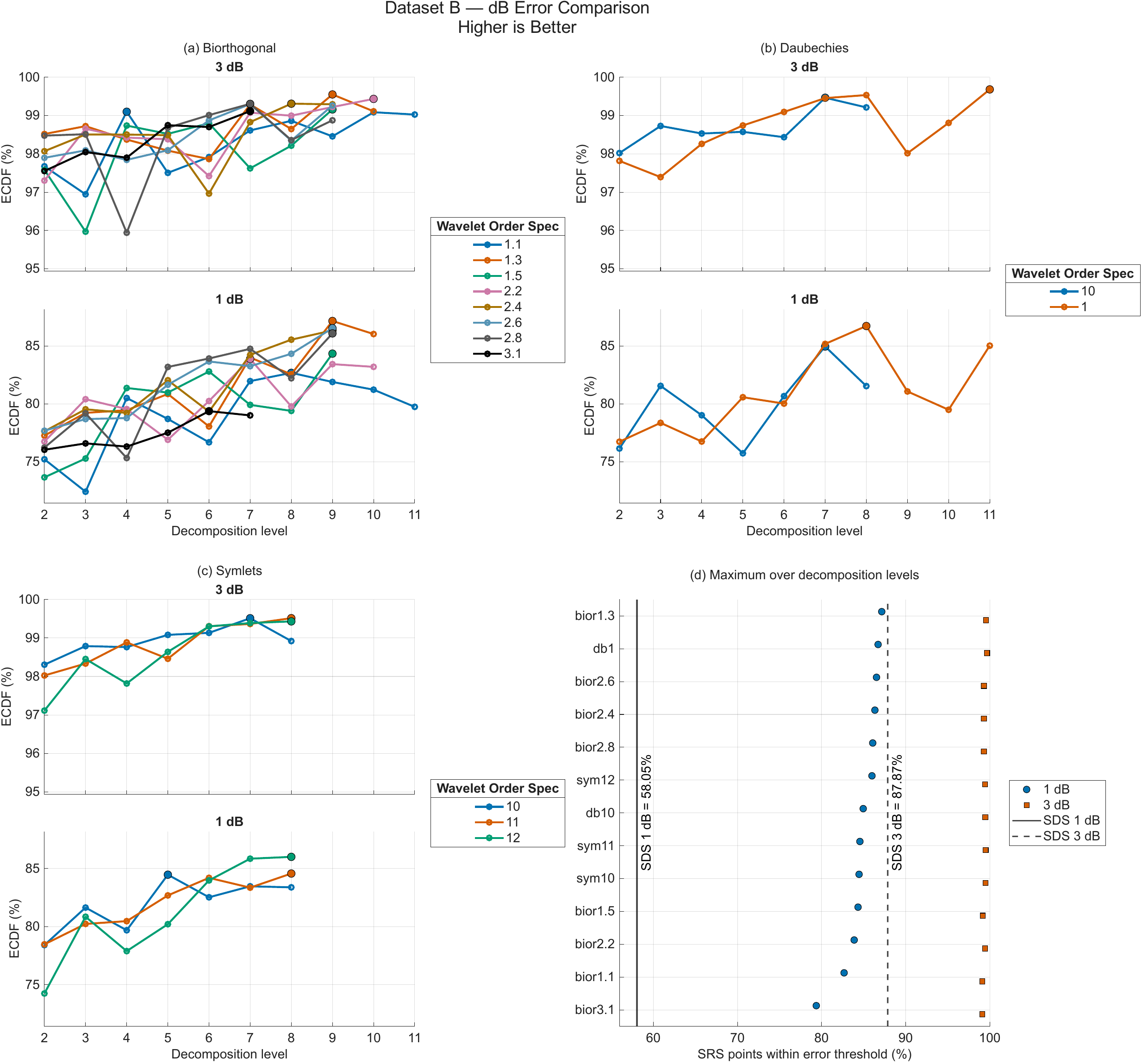}
    \caption{Percentage of pooled SRS points within 1~dB and 3~dB error thresholds by wavelet configuration for Dataset B.}
    \label{fig:wavelet-comparison-b-dB}
\end{figure}

\begin{figure}[H]
    \centering
    \includegraphics[width=\linewidth]{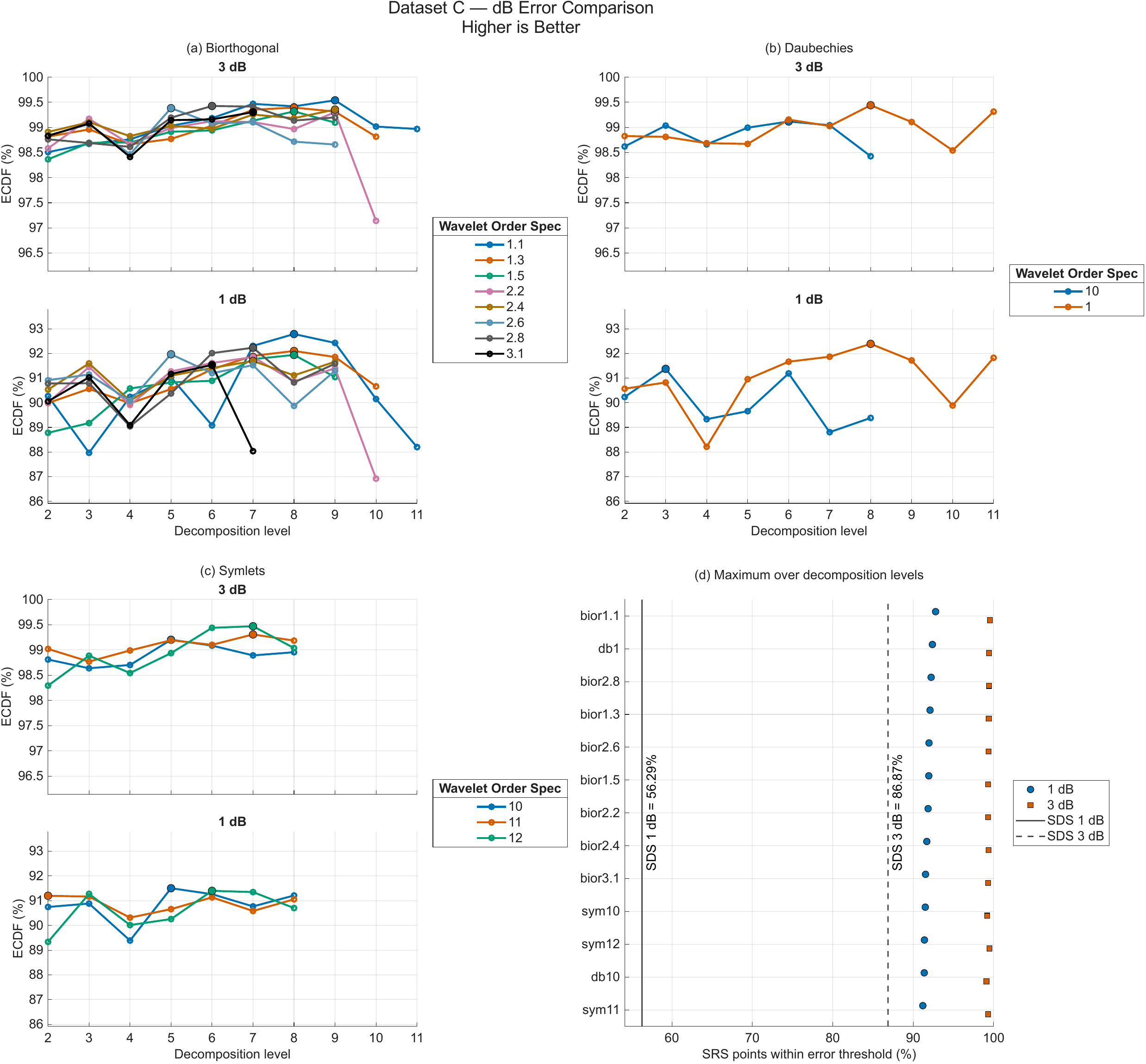}
    \caption{Percentage of pooled SRS points within 1~dB and 3~dB error thresholds by wavelet configuration for Dataset C.}
    \label{fig:wavelet-comparison-c-dB}
\end{figure}

\begin{figure}[H]
    \centering
    \includegraphics[width=\linewidth]{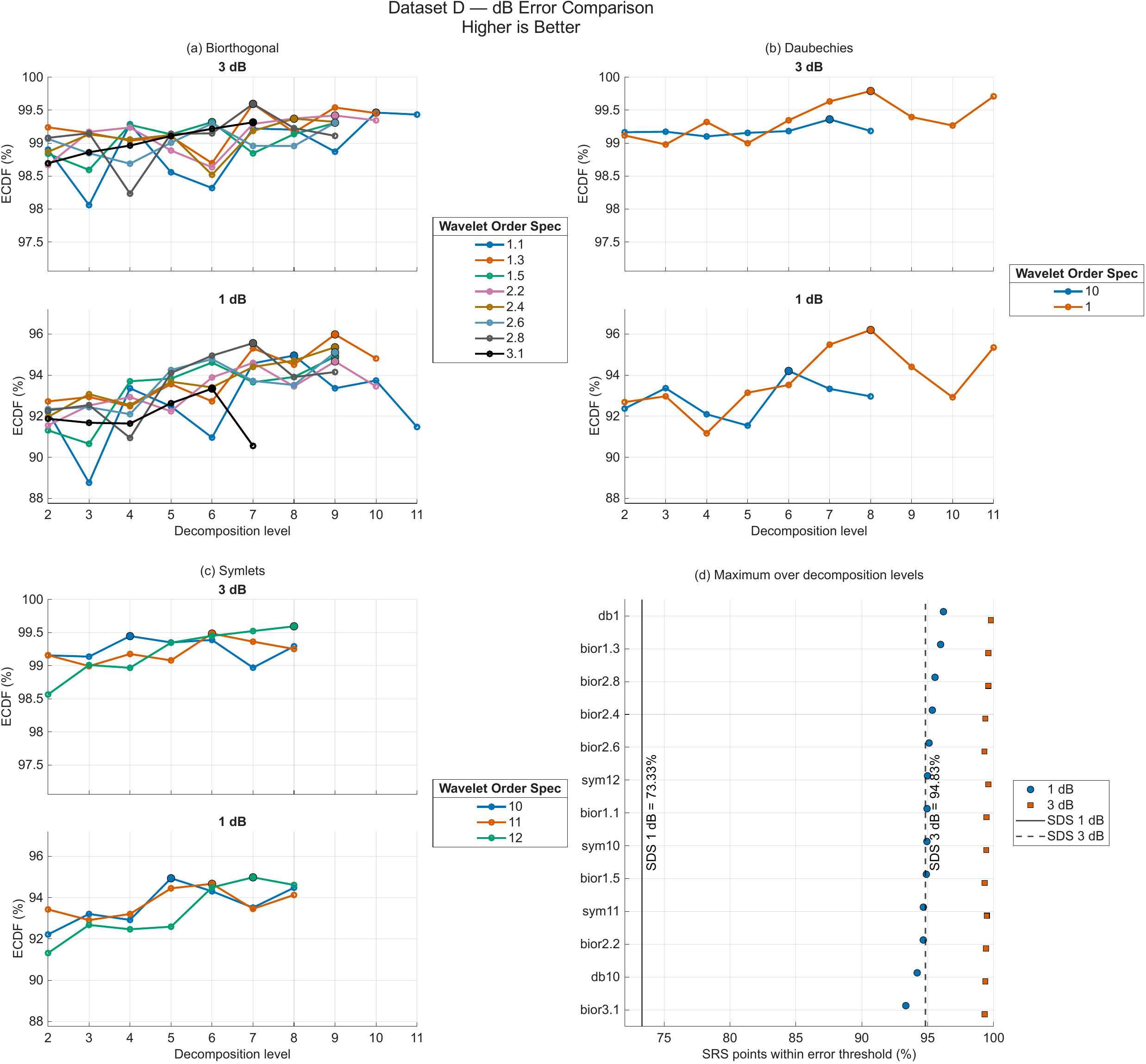}
    \caption{Percentage of pooled SRS points within 1~dB and 3~dB error thresholds by wavelet configuration for Dataset D.}
    \label{fig:wavelet-comparison-d-dB}
\end{figure}

Figures~\ref{fig:wavelet-comparison-a-dB}--\ref{fig:wavelet-comparison-d-dB} summarize the percentages of pooled SRS points with absolute errors at or below 1~dB and 3~dB for Datasets A--D, respectively. Errors are pooled across benchmark samples and SRS evaluation natural frequencies. In each figure, panels (a)--(c) show performance versus decomposition level for biorthogonal, Daubechies, and Symlet wavelets, respectively, with the 3~dB threshold above the 1~dB threshold. For each dataset, the CVAE ECDF values at 1~dB and 3~dB in Figure~\ref{fig:Holdout_dbEror_ABCD}, expressed as percentages, correspond to the \texttt{db10}, level~8 points for the respective thresholds in panel~(b) of that dataset's wavelet-comparison figure.

For the fixed 9000-sample signal length, maximum candidate decomposition levels vary with each wavelet's filter length, as described in \ref{sec:Multi-Model Wavelet Ensemble}. Not all eligible configurations were trained because of computational cost; the plotted points represent configurations that were trained and evaluated. Filled markers indicate the highest observed percentage for each wavelet and threshold, including ties. Panel~(d) shows the maximum percentage across the tested levels for each wavelet, with rows ordered by decreasing 1~dB performance. The maxima are computed independently for the two thresholds and may therefore correspond to different decomposition levels. Labeled vertical lines indicate the SDS baseline at each threshold. Higher percentages indicate better performance.

\FloatBarrier

\FloatBarrier
\section{Detailed SRS Comparisons with Brake (2011)}
\label{app:brake-comparisons}

Tables~\ref{tab:brake-table-1-comparison}--%
\ref{tab:brake-table-4-comparison} compare CVAE reconstructions
with the results reported by Brake~\cite{brake_inverse_2011}.
All comparisons are restricted to 10~Hz--10~kHz, excluding
the 20~kHz ordinate from each original target spectrum.
All SRS ordinates are expressed in $\mathrm{m/s^2}$.
Error metrics are evaluated over the natural frequencies listed
in each table, with the lowest error in each row shown in bold.

\begin{table}[htbp]
\centering
\caption{SRS comparison for the target in Table~1 of Brake~\cite{brake_inverse_2011}.}
\label{tab:brake-table-1-comparison}
\small
\setlength{\tabcolsep}{5pt}
\begin{tabular}{@{}rrrr@{}}
\toprule
& & \multicolumn{1}{c}{Brake (2011)} & \shortstack{Present study} \\
\cmidrule(lr){3-3}
\shortstack{Natural frequency\\(Hz)} & Target & Original & CVAE \\
\midrule
10 & 2 & 1.960 & 2.150 \\
100 & 6 & 5.709 & 6.349 \\
400 & 6 & 5.937 & 6.272 \\
500 & 9 & 9.350 & 9.087 \\
600 & 9 & 7.484 & 8.879 \\
700 & 6 & 6.309 & 6.043 \\
2000 & 8 & 8.044 & 7.712 \\
2500 & 11 & 10.270 & 10.801 \\
3000 & 8 & 8.289 & 7.929 \\
5000 & 8 & 8.400 & 7.993 \\
10000 & 9 & 8.831 & 9.089 \\
\midrule
\multicolumn{2}{@{}l}{MAE (dB)} & 0.419 & \textbf{0.219} \\
\multicolumn{2}{@{}l}{RMSE (dB)} & 0.584 & \textbf{0.294} \\
\multicolumn{2}{@{}l}{Maximum absolute error (dB)} & 1.602 & \textbf{0.628} \\
\bottomrule
\end{tabular}
\end{table}

\begin{table}[htbp]
\centering
\caption{SRS comparison for the target in Table~2 of Brake~\cite{brake_inverse_2011}.}
\label{tab:brake-table-2-comparison}
\small
\setlength{\tabcolsep}{5pt}
\begin{tabular}{@{}rrrr@{}}
\toprule
& & \multicolumn{1}{c}{Brake (2011)} & \shortstack{Present study} \\
\cmidrule(lr){3-3}
\shortstack{Natural frequency\\(Hz)} & Target & Original & CVAE \\
\midrule
10 & 10 & 10.030 & 9.927 \\
100 & 11 & 11.050 & 11.115 \\
1000 & 16 & 15.800 & 16.121 \\
1500 & 15 & 15.060 & 15.053 \\
10000 & 15 & 15.100 & 15.656 \\
\midrule
\multicolumn{2}{@{}l}{MAE (dB)} & \textbf{0.053} & 0.124 \\
\multicolumn{2}{@{}l}{RMSE (dB)} & \textbf{0.061} & 0.176 \\
\multicolumn{2}{@{}l}{Maximum absolute error (dB)} & \textbf{0.109} & 0.372 \\
\bottomrule
\end{tabular}
\end{table}

\begin{table}[htbp]
\centering
\caption{SRS comparison for the target in Table~3 of Brake~\cite{brake_inverse_2011}.}
\label{tab:brake-table-3-comparison}
\small
\setlength{\tabcolsep}{5pt}
\begin{tabular}{@{}rrrrr@{}}
\toprule
& & \multicolumn{2}{c}{Brake (2011)} & \shortstack{Present study} \\
\cmidrule(lr){3-4}
\shortstack{Natural frequency\\(Hz)} & Target & Original & GA-refined & CVAE \\
\midrule
10 & 5 & 5.019 & 5.033 & 4.739 \\
100 & 5 & 6.355 & 5.024 & 5.000 \\
200 & 7 & 7.340 & 7.543 & 7.007 \\
500 & 7 & 8.589 & 7.258 & 7.167 \\
1000 & 9 & 9.929 & 8.192 & 8.912 \\
1500 & 10 & 10.220 & 10.250 & 10.434 \\
2000 & 10 & 10.440 & 12.000 & 10.070 \\
3000 & 15 & 14.960 & 15.000 & 15.421 \\
5000 & 15 & 15.070 & 15.080 & 15.515 \\
8000 & 17 & 17.050 & 16.680 & 17.215 \\
10000 & 14 & 13.950 & 14.090 & 14.556 \\
\midrule
\multicolumn{2}{@{}l}{MAE (dB)} & 0.531 & 0.359 & \textbf{0.198} \\
\multicolumn{2}{@{}l}{RMSE (dB)} & 0.883 & 0.586 & \textbf{0.248} \\
\multicolumn{2}{@{}l}{Maximum absolute error (dB)} & 2.083 & 1.584 & \textbf{0.465} \\
\bottomrule
\end{tabular}
\end{table}

\begin{table}[htbp]
\centering
\caption{SRS comparison for the target in Table~4 of Brake~\cite{brake_inverse_2011}.}
\label{tab:brake-table-4-comparison}
\small
\setlength{\tabcolsep}{5pt}
\begin{tabular}{@{}rrrrr@{}}
\toprule
& & \multicolumn{2}{c}{Brake (2011)} & \shortstack{Present study} \\
\cmidrule(lr){3-4}
\shortstack{Natural frequency\\(Hz)} & Target & Original & GA-refined & CVAE \\
\midrule
10 & 8 & 8.089 & 8.076 & 7.532 \\
100 & 10 & 11.240 & 10.010 & 10.560 \\
200 & 12 & 12.370 & 11.790 & 12.012 \\
500 & 13 & 13.020 & 12.910 & 13.219 \\
1000 & 13 & 13.020 & 13.010 & 12.760 \\
1500 & 15 & 14.780 & 15.000 & 14.517 \\
2000 & 15 & 15.040 & 15.050 & 14.938 \\
3000 & 17 & 17.030 & 16.960 & 17.214 \\
5000 & 15 & 14.480 & 15.030 & 15.561 \\
8000 & 14 & 14.750 & 14.300 & 14.457 \\
10000 & 14 & 14.960 & 14.050 & 14.409 \\
\midrule
\multicolumn{2}{@{}l}{MAE (dB)} & 0.264 & \textbf{0.054} & 0.236 \\
\multicolumn{2}{@{}l}{RMSE (dB)} & 0.400 & \textbf{0.080} & 0.283 \\
\multicolumn{2}{@{}l}{Maximum absolute error (dB)} & 1.015 & \textbf{0.184} & 0.524 \\
\bottomrule
\end{tabular}
\end{table}
\FloatBarrier

\section*{Data availability}
 The code for synthetic waveform generation and SRS evaluation, along with the previously introduced benchmark datasets, is publicly available at \url{https://github.com/lanl/ML-Shock-Time-Series-Synthesis}. Model architectures and trained weights are available from the corresponding author upon reasonable request.

\bibliographystyle{elsarticle-num}
\bibliography{CVAE_Ensembles}

\end{document}